\documentclass{article}

\usepackage{url}

\usepackage{hyperref}% http://ctan.org/pkg/hyperref
\hypersetup{
  colorlinks=true,
	citecolor=blue!50!blue,
  linkcolor=blue!50!blue,
  urlcolor=blue!70!blue
}

\usepackage{times}
\usepackage{rotating}
\usepackage{multirow}
\usepackage{natbib}
\usepackage{amssymb}
\usepackage{amsmath}
\usepackage{mathtools}
\usepackage{multirow}
\usepackage{tcolorbox}
\usepackage{tabularx}
\usepackage{array}
\usepackage{colortbl}
\usepackage{fancybox}
\usepackage{tikz}

\usepackage{rotating}

\tcbuselibrary{skins}

\usepackage{pifont}% http://ctan.org/pkg/pifont
\usepackage[T1]{fontenc}    % use 8-bit T1 fonts
\usepackage{hyperref}       % hyperlinks
\usepackage{url}            % simple URL typesetting

\usepackage{amsfonts}       % blackboard math symbols
\usepackage{nicefrac}       % compact symbols for 1/2, etc.
\usepackage{microtype}      % microtypography
\usepackage{amssymb}
\usepackage{amsmath}
\usepackage{mathtools}
\usepackage{bigdelim} 
\usepackage{caption}

\usepackage{color}
\usepackage{verbatim}
\usepackage{siunitx}

\tcbuselibrary{skins}
\usepackage{fancybox}
\usepackage{tikz}
\usepackage{multirow}
\usepackage{tcolorbox}
\usepackage{tabularx}
\usepackage{array}
\usepackage{colortbl}
\usepackage{epstopdf}

\usepackage{hyperref}

\usepackage[font=small,skip=5.5pt]{caption}

\usepackage[utf8]{inputenc} % allow utf-8 input
\usepackage[T1]{fontenc}    % use 8-bit T1 fonts
\usepackage{hyperref}       % hyperlinks
\usepackage{url}            % simple URL typesetting
\usepackage{booktabs}       % professional-quality tables
\usepackage{amsfonts}       % blackboard math symbols
\usepackage{nicefrac}       % compact symbols for 1/2, etc.
\usepackage{microtype}      % microtypography

\usepackage{color}
\usepackage{enumitem}

\usepackage{calc}

\newlength{\spacer}
\newsavebox{\mybox}

\newcommand*{\xdash}[1][1.5em]{\rule[0.25ex]{#1}{1.55pt}}

\newcommand*{\xdashed}[1][.25em]{\rule[0.25ex]{#1}{1.55pt}}

\newcommand*{\rxdashed}{{\color{red}{\xdashed}\,\color{red}{\xdashed}\,\color{red}{\xdashed}\,\color{red}{\xdashed}\,\color{red}{\xdashed}}}
\newcommand*{\rxdash}{{\color{red}{\xdash}}}

\usepackage{microtype}
\usepackage{graphicx}
\usepackage{subfigure}
\usepackage{booktabs} % for professional tables

\usepackage{hyperref}

\usepackage[accepted]{icml2020}

\icmltitlerunning{Frequentist Uncertainty in RNNs via Blockwise Influence Functions}

\begin{document}

\twocolumn[
\icmltitle{Frequentist Uncertainty in Recurrent Neural Networks via\\ \textit{Blockwise} Influence Functions}




\begin{icmlauthorlist}
\icmlauthor{Ahmed M. Alaa}{to}
\icmlauthor{Mihaela van der Schaar}{to,too}
\end{icmlauthorlist}

\icmlaffiliation{to}{University of California, Los Angeles}
\icmlaffiliation{too}{Cambridge University}

\icmlcorrespondingauthor{Ahmed M. Alaa}{ahmedmalaa@ucla.edu}

\icmlkeywords{Machine Learning, ICML}

\vskip 0.3in
]



\printAffiliationsAndNotice{}  % leave blank if no need to mention equal contribution

\begin{abstract} 
Recurrent neural networks (RNNs) are instrumental in modelling sequential and time-series~data. Yet, when using RNNs to inform decision-making, predictions by themselves are not~sufficient~---~we also need estimates of predictive {\it uncertainty}. Existing approaches for uncertainty~quantification~in RNNs are based predominantly on Bayesian methods; these are computationally prohibitive, and require major alterations to the RNN architecture and training. Capitalizing on ideas from classical jackknife resampling, we develop a {\it frequentist}~alternative that: (a) does not interfere with model training or compromise its accuracy, (b) applies to {\it any} RNN architecture, and (c) provides theoretical coverage guarantees on the estimated uncertainty intervals. Our method derives predictive uncertainty from the variability of the (jackknife) sampling distribution of the RNN outputs, which is estimated by repeatedly deleting ``blocks'' of (temporally-correlated) training data, and collecting the predictions of the RNN re-trained on the remaining data. To avoid exhaustive re-training, we utilize {\it influence functions} to estimate the effect of removing training data blocks on the learned RNN parameters. Using data from a critical care setting, we demonstrate the utility of uncertainty quantification in sequential decision-making.
\end{abstract}

\section{Introduction}
\label{Sec1} 
Recurrent neural networks (RNNs) are sequence-based models of key importance in a wide range of application domains \cite{sundermeyer2012lstm,kalchbrenner2013recurrent}. 
Because of their ability to handle~temporal~sequences,~RNNs have been used in applications that involve sequential decision making over time --- examples of such applications include: stock market predictions \cite{bao2017deep,selvin2017stock}, service demand forecasting \cite{wen2017multi,zhu2017deep}, medical prognoses \cite{shickel2017deep, lim2018forecasting, alaa2019attentive}~and~reinforcement learning \cite{bakker2002reinforcement,wang2018supervised}. 

\begin{figure}[t]
  \centering
  \includegraphics[width=3.25in]{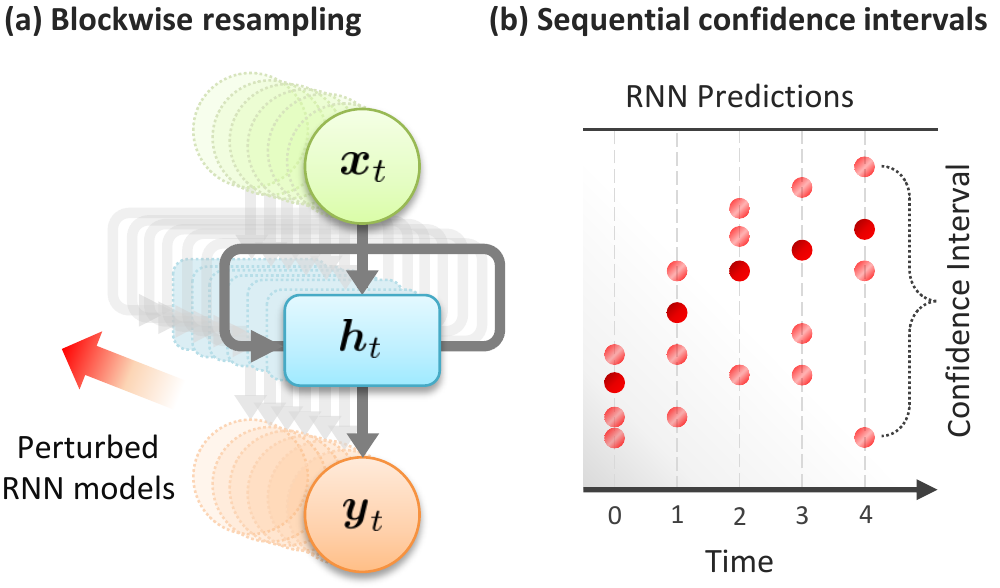}
  \caption{\footnotesize {\bf Pictorial depiction of our approach.}~(a)~Blockwise~resampling is applied to the RNN in a {\it post-hoc} fashion to create~$N$ perturbed versions of the model $\{\mbox{RNN}_1,\ldots,\mbox{RNN}_N\}$ (illustrated through the shaded diagrams) by repeatedly deleting $N$ ``blocks'' of training data, and using {\it influence functions} to estimate the parameters of the perturbed models. Here, $h_t$ is the RNN's hidden state, $x_t$ is its input and $y_t$ is its output, (b) Confidence intervals are constructed on the basis of the variability of the $N$ perturbed RNNs' outputs (highlighted in light red color). Here, we set $N=3$.} 
  \rule{\linewidth}{.75pt}  
  \label{Fig1} 
\vspace{-9.75mm}  
\end{figure}

In high-stakes applications where RNNs would inform critical decisions, reliable estimates of {\it predictive uncertainty} are indispensable for accurate risk assessment \cite{gollier2018economics}. For instance, optimal decisions on {\it when} to administer an aggressive treatment to a hospitalized patient (based on their temporal lab measurements) requires estimating the uncertainty in their predicted outcomes in order to properly weigh risks against benefits \cite{dusenberry2019analyzing}. Whereas various methods for uncertainty estimation in feed-forward networks have been recently proposed \cite{lakshminarayanan2017simple,malinin2018predictive,maddox2019simple}, equivalent methods for RNN models are still lacking. 

In this paper, we develop a principled {\it frequentist} approach for estimating uncertainty in the predictions of RNN models. (A high-level depiction of our approach is given in~Figure~\ref{Fig1}.) Our method builds~on~classical~{\it jackknife}~resampling~\cite{miller1974jackknife} to create multiple ``perturbed'' versions of an RNN by resampling its trained parameters (Figure \ref{Fig1}.a). Through the RNN's forward pass, we estimate the sampling distribution of the model's predictions by collecting~the outputs of its perturbed copies --- confidence intervals~on the model's~predictions are then constructed based on the jackknife residuals and the variability of the resampled outputs (Figure\ref{Fig1}.b). The entire procedure is applied in a {\it post-hoc} fashion, hence it neither interferes with model training nor compromises its accuracy. Moreover, since the jackknife is model-agnostic, our approach covers a wide range of RNN architectures.

Because RNNs process (temporally-correlated) sequences, our jackknife resampling scheme operates on ``blocks''~of~dependent data points instead of individual observations.~That is, in order to resample the parameters $\hat{\boldsymbol{\theta}}_{\mbox{\scriptsize RNN}}$ of an RNN trained on a data set $\mathcal{D}$, we repeatedly delete temporally~contiguous blocks of data from $\mathcal{D}$, and estimate the parameters of the perturbed version of the model --- trained on~the~remaining data --- through the following relationship: 
\begin{align}
\mathcal{I}(\mbox{Data Block}) \approx \hat{\boldsymbol{\theta}}_{\mbox{\scriptsize RNN}}(\mathcal{D}\setminus\mbox{Data Block}) - \hat{\boldsymbol{\theta}}_{\mbox{\scriptsize RNN}}(\mathcal{D}), \nonumber
\end{align}
where $\mathcal{I}(.)$ is a {\it blockwise influence function} that quantifies the effect of removing a data block in $\mathcal{D}$ on the~learned~RNN parameters --- this is a generalization of conventional influence functions defined with respect to individual~data~points \cite{hampel2011robust,koh2017understanding}. In~Section~\ref{Sec4}, we develop a stochastic estimation algorithm, based on blockwise batch sampling, to compute $\mathcal{I}(.)$ in linear time for RNN models trained via backpropagation through time (BPTT) or any of its variants. Conducting our blockwise resampling scheme requires only access to the model's loss gradients, and is independent of the model's specific architecture. 

Using the ``leave-one-block-out'' training residuals and the variability of the RNN predictions --- estimated via influence functions --- we construct a sequence of confidence intervals that ``sandwiches'' the sequential predictions made by the RNN model. In Section \ref{Sec3}, we extend the classical notion of {\it coverage probability} of confidence intervals, i.e., the likelihood that a confidence interval contains the true label \cite{marra2012coverage}, to the sequential prediction setup, and in Section \ref{Sec4}, we show that under certain conditions, confidence intervals estimated through blockwise resampling can achieve any pre-specified coverage probability.

Finally, in Section \ref{Sec5}, we~demonstrate~the~utility~of~our proposed method in the context of sequential decision-making in critical care medicine using real-world data, in addition to synthetic simulations. More specifically, we show that uncertainty estimates generated by our method (on top of an RNN model) can inform clinicians on {\it when} antibiotics can be {\it safely} prescribed to hospitalized critically-ill patients so that improvements in their health outcome are maximized while taking into account uncertainties about side effects.  

\newpage
\section{Related Work}
\label{Sec2} %Influence functions have been
Driven by the desire to suppress the ``overconfidence'' exhibited by poorly calibrated --- but highly discriminative --- deep learning models, various methods for uncertainty estimation in standard feed-forward neural networks have been recently proposed \cite{hernandez2015probabilistic,gal2016dropout,lakshminarayanan2017simple,malinin2018predictive}. However, the literature on uncertainty quantification in RNNs is rather scarce. In what follows, we provide a brief glimpse at three strands of relevant literature; a more elaborate discussion on these methods, the history of jackknife resampling, and the connections between our work and modern applications of influence functions \cite{koh2017understanding,koh2019accuracy} is provided in Appendix~A.

{\bf Bayesian RNNs.} The most prevalent approaches for RNN uncertainty estimation hinge on Bayesian modeling \cite{fortunato2017bayesian, chien2015bayesian, zhu2017deep,mirikitani2009recursive}. By~specifying~a~prior~over~the RNN parameters, these~approaches~estimate a model's uncertainty via its posterior credible intervals. However, exact inference in Bayesian deep learning is generally intractable or computationally infeasible. Hence, practical Bayesian RNNs rely on dropout-based approximate inference \cite{gal2016dropout, gal2016theoretically}, which is often hard to calibrate, and is generally ill-posed as its posterior distributions do not concentrate asymptotically \cite{osband2016risk,hron2017variational}. Moreover, Bayesian methods require significant modifications to model architecture and training which limits their versatility. Our paper addresses these issues by developing a frequentist post-hoc alternative to Bayesian methods.

{\bf Quantile RNNs.} Similar to quantile regression \cite{koenker2001quantile}, quantile RNNs (Q-RNNs) learn prediction intervals by explicitly modeling the cumulative distribution function of the prediction targets \cite{taylor2000quantile, wen2017multi,gasthaus2019probabilistic}. The key difference between these approaches and ours is that Q-RNNs learn uncertainty jointly with the main prediction task, which as we show in Section \ref{Sec5}, can compromise their predictive accuracy. Moreover, Q-RNNs are bespoke models that must be retailored for each new prediction task, whereas our approach can be applied in a post-hoc fashion to any RNN architecture.  Empirical comparisons between our method and the 2 approaches discussed above are conducted in Section \ref{Sec5}. 

{\bf Probabilistic RNNs.} This class of models capture the structural uncertainty within the stochastic dynamics of sequential data. Existing variants of probabilistic RNNs are mainly based on deep state-space models \cite{krishnan2017structured, rangapuram2018deep,alaa2019attentive}. Probabilistic models are bespoke to the application at hand, and may not be appropriate for some application domains. Moreover, most probabilistic models entail restrictive assumptions on sequence dynamics (such as Markovianity). %Empirical comparisons between our method and the 3 approaches discussed above are conducted in Section \ref{Sec5}.   
%Jackknife resampling, originally developed by Quenouille in \cite{quenouille1956notes} and refined by Tukey~in~\cite{tukey1958bias},~is a general methodology for estimating sampling distributions. Our work builds on this literature by applying~the jackknife to RNN predictions in order to estimate its predictive uncertainty through the sampling variance. The key difference between our setup and that of the standard jackknife is that RNNs do not operate on i.i.d data --- they operate on temporally correlated sequences. In \cite{kunsch1989jackknife},~K\"unsch proposed a variant of the jackknife for general stationary observations rather than i.i.d data. As we show in Section~\ref{Sec5}, the sampling procedure therein is a special case of ours~when the RNN is trained with a single sequence. 
  
\newpage
\section{Frequentist Uncertainty in RNNs}
\label{Sec3} 
This Section lays out the class of sequential prediction~tasks and RNN~architectures~under~study~(Section~\ref{Sec31}),~provides a formal description~of~sequential~data~(Section~\ref{Sec32}), and extends classical notions of frequentist coverage to the sequential setup under consideration (Section \ref{Sec33}).

\subsection{Sequential Prediction with RNNs}
\label{Sec31}
We consider a broad class of sequential prediction~tasks,~to which we provide an encapsulating formulation as follows. Given a (temporal) input sequence $\boldsymbol{x} = [x_1,\ldots, x_{T}]$ of length $T$, our objective is to predict a corresponding output sequence $\boldsymbol{y} = [y_1,\ldots, y_{T+T^{\prime}}]$, where $T^{\prime} \geq 0$ is the prediction horizon, i.e., the number of future time steps predicted at each time step $t$. To this end, an RNN is trained to learn the conditional expectation of future values of $\boldsymbol{y}$, i.e.,
\begin{align}
\mbox{RNN}(x_1,\ldots,x_t;\boldsymbol{\theta}) \,\approx\, \mathbb{E}[\,\boldsymbol{y}_{t:t+T^{\prime}}\,|\,\boldsymbol{x}_{1:t}, \boldsymbol{y}_{1:t}\,], 
\label{Eq1}
\end{align}
where $\boldsymbol{x}_{t:t^{\prime}} = [x_t,\ldots,x_{t^{\prime}}]$, $\boldsymbol{\theta}$ is the RNN model parameters, and the expectation $\mathbb{E}[\,.\,]$ is taken with respect to the randomness of the stochastic process $\mathcal{S} = (\boldsymbol{x}, \boldsymbol{y})$, which we thoroughly describe in Section \ref{Sec32}. Throughout the paper, we abstract away the RNN model through a sequence of functions $\{f_t\}^T_{t=1}$, $f_t: \mathcal{X}^t \to \mathcal{Y}^{T^{\prime}}$, where $\mathcal{X}$ and $\mathcal{Y}$ are the input and output spaces, respectively. The RNN prediction~for the $t^{\prime}$-th horizon at time $t$ is denoted by $f_{t,t^{\prime}}(\boldsymbol{x}_{1:t}; \boldsymbol{\theta})$.

{\bf Model training.} A training data set $\mathcal{D}_n = \mbox{\small $\{\mathcal{S}^{(i)}, 1 \leq i \leq n\}$}$ comprises a set of $n$ sequences\footnote{Our formulation and proposed methods can be straightforwardly extended to accommodate variable-length sequences.}, where \mbox{\small $\mathcal{S}^{(i)} = (\boldsymbol{x}^{(i)}, \boldsymbol{y}^{(i)})$} is the $i$-th sequence. The model is trained by minimizing the loss function \mbox{\small $L(\mathcal{D}_n;\boldsymbol{\theta},\boldsymbol{w}) = \sum_{i=1}^{n} L(\mathcal{S}^{(i)};\boldsymbol{\theta},\boldsymbol{w})$}, where 
\begin{align}
L(\mathcal{S}^{(i)};\boldsymbol{\theta},\boldsymbol{w}) \triangleq \sum_{t,t^{\prime}}\,w^{(i)}_{t,t^{\prime}}\cdot\ell(y^{(i)}_{t+t^{\prime}}, f_{t,t^{\prime}}(\boldsymbol{x}^{(i)}_{1:t}; \boldsymbol{\theta})).
\label{Eq2}
\end{align}
That is, the model's loss at the $t^{\prime}$-th horizon in the $t$-th time step for the $i$-th sequence is weighted by $w^{(i)}_{t,t^{\prime}} \in \mathbb{R}_+$. We represent the weights $\{w^{(i)}_{t,t^{\prime}}:\, \forall i,t,t^{\prime}\}$ via the (flattened) vector $\boldsymbol{w} \in \mathbb{R}^{n T T^{\prime}}$. The choice of the~loss~$\ell(.)$~depends on whether $\boldsymbol{y}$ is a regression or a classification target. 

{\bf Which RNN architectures does our formulation cover?} For different settings of the weight vector $\boldsymbol{w}$ and the prediction horizon $T^{\prime}$, the general formulation in (\ref{Eq2}) encapsulates many prediction setups and RNN architectures common in various applications. A non-exhaustive list of examples for such architectures are provided in Table \ref{Table1} --- these include the {\it many-to-many} architecture used in sequence labeling tasks \cite{luong2015multi}, the {\it many-to-one} architecture used in sequence classification tasks \cite{graves2006connectionist}, and the {\it sequence-to-sequence} architecture, usually implemented via encoder-decoder schemes \cite{sutskever2014sequence}. 

\begin{table}[h] %[htbp]
\centering
\caption{Generality of the formulation in (\ref{Eq2}).}
\vspace{0.05in}
\begin{tabular}{ll} %{ |c|c| }
%\hline
\toprule
\mbox{\footnotesize {\bf RNN model}} & \mbox{\footnotesize {\bf Weight vector $\boldsymbol{w}$}} \\[2px]  
\hline
&\\[-5px]
\mbox{\footnotesize {Sequence labeling}} & \mbox{\footnotesize $T^{\prime}=0, \,w^{(i)}_{t,0} = 1,\, \forall\, i, t$.}\\[2px] 
\mbox{\footnotesize {Sequence classification}} & \mbox{\footnotesize $T^{\prime}=0,\, w^{(i)}_{t,0} = \boldsymbol{1}_{\{t=T\}},\, \forall\, i$.}\\[2px] 
\mbox{\footnotesize {Sequence-to-sequence}} & \mbox{\footnotesize $T^{\prime}>0,\, w^{(i)}_{t,t^{\prime}} = \boldsymbol{1}_{\{t=T\}},\, \forall\, i, t^{\prime}$}.\\[5px] 
\bottomrule 
\end{tabular}
\label{Table1}
\vspace{-4mm} 
\end{table}
  
We do not pose any assumptions on the type of representations learned by the RNN --- our proposed methods apply to LSTM, GRU and attention-based architectures. While our work is primarily motivated by applications in which the sequence $\mathcal{S}$ is {\it temporal}, all methods presented in Section~\ref{Sec4} are applicable to other sequential setups, such as natural language processing, where sequences are not necessarily defined over time \cite{auger2008expression}. 

\begin{figure*}[t]
  \centering
  \includegraphics[width=6.25in]{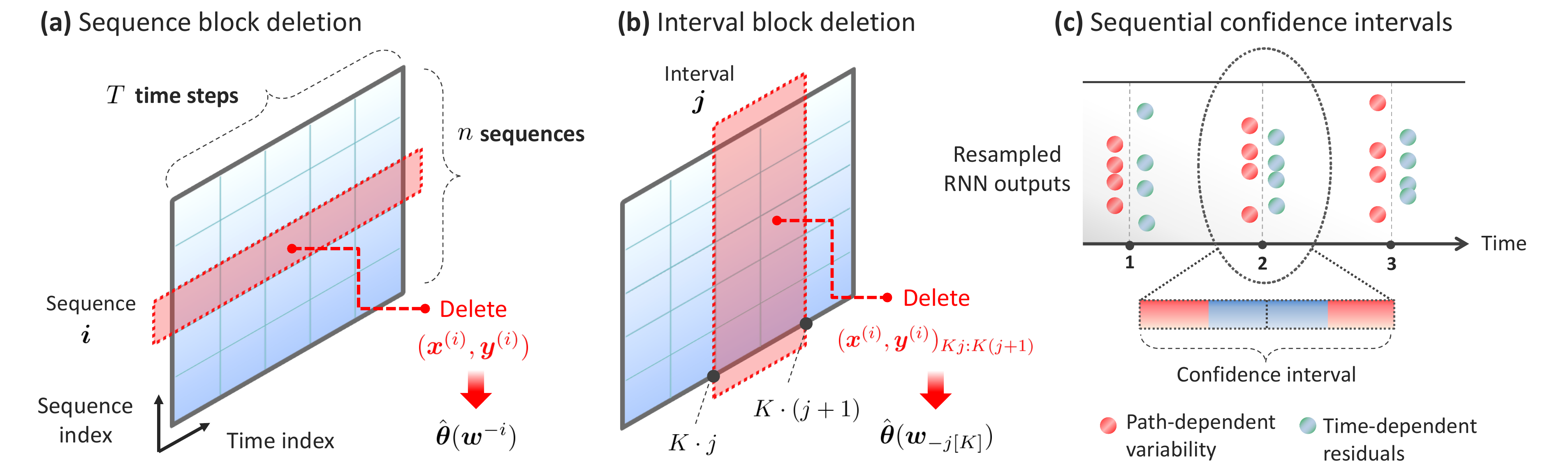}
  \caption{\footnotesize {\bf Uncertainty estimation in RNNs via BJ resampling.} (a) {\it Sequence block deletion.} In this panel, we represent the data~set~$\mathcal{D}_n$ as an $n \times T$ matrix. Sequence deletion is conducted by iteratively deleting rows from this matrix. (b) {\it Interval block deletion.} The columns of the matrix $\mathcal{D}_n$ are grouped into $\lfloor T/K \rfloor$ partitions, each spanning $K$ columns --- interval deletion proceeds by deleting one partition at a time. The data set $\mathcal{D}_n$ is resampled $n \cdot \lfloor T/K \rfloor$ times by repeatedly applying sequence and interval block deletion in a nested fashion as shown in Algorithm 1. (c) {\it Confidence intervals construction.} By re-training the RNN model once on each resampled data set, we collect multiple perturbed versions of the trained parameters $\hat{\boldsymbol{\theta}}$. For a new test sequence, confidence intervals are determined by the quantiles of the leave-one-block-out residuals (red), which only depend on the training data and are fixed for each time step, in addition to the predictions made by the perturbed models for the test sequence (blue), which account for path-dependent variability of the RNN outputs.} 
  \rule{\linewidth}{.75pt}   
  \label{Fig2} 
  \vspace{-7mm} 
\end{figure*}
  
\subsection{Sequential Data as Stochastic Processes}
\label{Sec32}
Given a probability space $(\Omega, \mathcal{H}, \mathbb{P})$, we view the sequence $\mathcal{S} = (\boldsymbol{x}, \boldsymbol{y})$ as a {\it stochastic process} comprising a collection of random variables on $\Omega$ indexed by time steps $\{1,\ldots,T\}$. A filtration ${\mathcal{H}_t, 1 \leq t \leq T,}$ is a (weakly) increasing collection of $\sigma$-algebras on $\Omega$, and is always bounded above by $\mathcal{H}$, i.e., $\mathcal{H}_t \subseteq \mathcal{H}$. The filtration $\mathcal{H}_t$ comprises the information that the RNN model uses to make a prediction at the $t$-th time step, i.e., the history of the process up to time $t$. 

We characterize the stochastic process $\mathcal{S}$ via its {\it $\beta$-mixing} rate. Formally, the $\beta$-mixing coefficient for $\mathcal{S}$ generated~under the distribution $\mathbb{P}$ is defined as \cite{mcdonald2011estimating}
\begin{align}
\beta_T(\tau) \triangleq \sup_{t} \|\,\mathbb{P}_{1:t} \otimes \mathbb{P}_{t+\tau:T} - \mathbb{P}_{t,\tau}\,\|_{TV},
\label{Eq3}
\end{align}
where $\|.\|_{TV}$ is the total variation norm, $\mathbb{P}_{t:t^{\prime}}$ is the joint distribution of $(\boldsymbol{x}_{t:t^{\prime}},\boldsymbol{y}_{t:t^{\prime}})$, and $\mathbb{P}_{t,t^{\prime}}$ is the joint distribution of $(\boldsymbol{x}_{1:t},\boldsymbol{y}_{1:t})$ and $(\boldsymbol{x}_{t+t^{\prime}:T},\boldsymbol{y}_{t+t^{\prime}:T})$. The $\beta$-coefficient $\beta_T(\tau)$ measures the extent to which the trajectory of $\mathcal{S}$ after $\tau$ time steps in the future depends on its history --- smaller values for $\beta_T(\tau)$ mean that $\mathcal{S}$ forgets its past. When $\beta_T(\tau)=0$ for all $\tau$, then the data at all time steps are independent, whereas a non-zero $\beta_T(\tau)=0$ for all $\tau$ corresponds to sequences where samples across all time steps exhibit correlation. 
%As we show later in Section \ref{Sec5}, the quality of uncertainty estimates issued by our method will depend fundamentally on $\beta_T(\tau)$.

\subsection{Sequential Confidence Intervals}
\label{Sec33} 
Our key objective is to quantify the uncertainty~in~the~RNN's sequential predictions $\{f_t(\boldsymbol{x}_{1:t}; \boldsymbol{\theta})\}^T_{t=1}$. Following a formal frequentist approach to the problem, Section \ref{Sec4} develops a procedure which maps the filtrations $\{\mathcal{H}_t\}_t$ to~a~sequence~of predictive confidence intervals\footnote{For simplicity of exposition, we consider~one-dimensional~outputs; for which uncertainty is quantified via prediction intervals. If $\boldsymbol{y}$ is multidimensional, then $\mathcal{C}_t$ is a {\it confidence region}.} $\{\mathcal{C}_t(\mathcal{H}_t;\alpha)\}_t$ that contain the true output $\boldsymbol{y}$ with a {\it coverage probability} of $(1-\alpha)$.  

{\bf Time-dependent coverage.} Conventional notions of coverage pertain to ``static'' regression setups \cite{lawless2005frequentist}. In our sequential setup, confidence estimates are issued for each time step in a test sequence --- so how can we extend the notion of coverage to accommodate temporality? To capture the temporal aspect, we coin the notion of ``time-dependent coverage'' as follows. An $\alpha$-level time-dependent coverage is satisfied if for every $t$, confidence intervals cover true outcomes with probability $1-\alpha$, i.e.,
\begin{align}
\mathbb{P}(\varphi(y_{t+t^{\prime}}) \in \mathcal{C}_{t,t^{\prime}}(\mathcal{H}_t;\alpha)) \geq 1 - \alpha,\, \forall\, t, t^{\prime}.
\label{Eq4}
\end{align} 
Here, $\varphi(.)$ is a function that maps $y_t$ to the true class probabilities $\{1-p, p\}$ if $y_t$ is a binary classification target, and is an identity function if $y_t$ is a regression target. (\ref{Eq4}) is an appropriate design objective in applications where sensitive decisions are made repeatedly at each time step --- we provide an example for such application in Section \ref{Sec5}. 

{\bf Distinguishing two types of~uncertainty.}~We~note~that,~for classification outputs, i.e., $y_t \in \{0,1\}$,~we~differentiate~between the notions of calibration (or risk) and model uncertainty \cite{fedorov2013theory, osband2016risk}. Calibration refers to the accuracy of the model's point estimates of the class probabilities, whereas model uncertainty refers to the model's confidence in those point estimates. Calibration is also often referred to as {\it aleatoric uncertainty} \cite{gal2016uncertainty} --- it reflects the inherent noise uncertainty in the data itself, rather than the model's epistemic uncertainty.~Our~sought-after~confidence intervals $\{\mathcal{C}_t\}_t$ implicitly quantify both~types~of~uncertainty, but do not explicitly disentangle them,~i.e.,~poor~calibration would implicitly manifest through wider confidence intervals. Explicit assessment of model calibration can be carried out by computing Brier scores \cite{avati2018countdown}.

\section{The Blockwise Infinitesimal Jackknife}
\label{Sec4}
How can we construct confidence intervals~$\{\mathcal{C}_t\}_t$~that~satisfy the coverage constraints in (\ref{Eq4})?~In~this~Section,~we do so through a formal resampling procedure that estimates the variability in the {\it sampling distribution} of an RNN's output by resampling its trained parameters $\hat{\boldsymbol{\theta}}(\boldsymbol{w})$.~(Figure~\ref{Fig2}~provides a high-level illustration of our procedure.)

\subsection{Jackknifing via Block Deletion}
\label{Sec41}
Given a data set $\mathcal{D}_n$ and a weight vector $\boldsymbol{w}$, the trained RNN parameters, obtained by minimizing the loss in (\ref{Eq2}), are
\begin{align}
\hat{\boldsymbol{\theta}}(\boldsymbol{w}) = \arg \min_{\boldsymbol{\theta}}\, L(\mathcal{D}_n;\boldsymbol{\theta}, \boldsymbol{w}).
\end{align}
We use {\it jackknife} resampling to estimate the sampling distribution of $\hat{\boldsymbol{\theta}}(\boldsymbol{w})$. The jackknife entails repeatedly re-training the RNN on leave-one-out (LOO) subsets of the training data $\mathcal{D}_n$, with one {\it individual} data point deleted from $\mathcal{D}_n$ at a time \cite{miller1974jackknife,efron1992jackknife}. Unlike the standard regression setup, wherein all training data points are independent \cite{tukey1958bias}, here we deal with temporally-correlated data sequences. Hence, to preserve the temporal dependence in the data, we delete {\it blocks} of temporally contiguous data points instead of individual observations --- we denote this ``leave-one-block-out'' scheme as the {\it blockwise} jackknife. 

Two types of data blocks can be used in our blockwise jackknife (BJ) scheme. The first is the {\it sequence~block},~depicted in Figure \ref{Fig2}.a, which comprises the entirety of a single sequence $\mathcal{S}=(\boldsymbol{x},\boldsymbol{y})$ in the training data $\mathcal{D}_n$. The second is the {\it interval block}, depicted in Figure \ref{Fig2}.b, which comprises the set of all sub-sequences of length $K \leq T$ occupying the same time indexes across all sequences in $\mathcal{D}_n$.~If~we~think of $\mathcal{D}_n$ as an $n \times T$ matrix, then sequence and interval blocks represent horizontal and vertical partitions of $\mathcal{D}_n$, respectively. BJ operates by repeatedly deleting interval and/or sequence blocks, and training the RNN on the remaining~data. Block deletion is achieved by setting specific elements~of $\boldsymbol{w}$ to zero, leading to an altered weight vector $\boldsymbol{w}^\prime$ as follows:  

{\bf (a) Sequence block deletion.} Deletion of~the~$i$-th~sequence $\mathcal{S}^{(i)}$ in $\mathcal{D}_n$ entails setting the weights $w^{(i)}_{t,t^{\prime}} = 0,\,\forall t,t^{\prime}$. We denote the resulting weight vector as $\boldsymbol{w}^{-i}$. Re-training using $\boldsymbol{w}^{-i}$ is equivalent to training on the data set $\mathcal{D}_n/\{\mathcal{S}^{(i)}\}$.    

{\bf (b) Interval block deletion.} For interval blocks of length $K$, we divide the sequence length $T$ into $\lfloor T/K \rfloor$ equal-sized blocks. Deletion of the $j$-th interval block~entails~setting~the weights $w^{(i)}_{j^\prime,t^\prime} = 0,\,\forall i,t^\prime, K\cdot j \leq j^\prime \leq K\cdot (j+1)$.~We denote the corresponding weight vector as $\boldsymbol{w}_{-j[K]}$. When deleting an interval block, we stitch together the remaining segments of each sequence, preserving temporal order. 

Sequence blocks are more relevant in applications where we would have many (short) training sequences (e.g., medical time series data), whereas interval blocks are relevant to applications with few (long) sequences (e.g., financial data). As shown in Algorithm 1, BJ resampling generally proceeds by deleting sequence and interval blocks in a nested fashion (we denote the weight vector with the $i$-th sequence and the $j$-th interval deleted as $\boldsymbol{w}^{\mbox{\tiny $-i$}}_{\mbox{\tiny $-j[K]$}}$.) Using the RNN parameters re-trained on $\boldsymbol{w}^{\mbox{\tiny $-i$}}_{\mbox{\tiny $-j[K]$}}$, we collect the leave-one-block-out (LOBO) error residuals of the re-trained RNN models, i.e., 
\begin{align}
r^{(i,j)}_{t, t^{\prime}} = \big|\,y^{(i)}_{t + t^{\prime}} - f_{t,t^{\prime}}(\boldsymbol{x}^{(i)}_{1:t}; \hat{\boldsymbol{\theta}}(\boldsymbol{w}^{\mbox{\tiny $-i$}}_{\mbox{\tiny $-j[K]$}}))\,\big|,
\label{Eq6}
\end{align}
for all $t^\prime$ and $K\cdot j \leq t \leq K\cdot (j+1)$. That is, the LOBO residuals correspond to the residual errors of the RNN re-trained with the $i$-th sequence and the $j$-th interval deleted, and evaluated on the data block defined by the intersection between the two deleted blocks, i.e., the~intersection~between the red blocks in Figures \ref{Fig2}.a and \ref{Fig2}.b. In most practical applications, we would need to delete only either sequence or interval blocks, depending on the nature of the application at hand.%Overall, the BJ process entails re-training the model $n \cdot \lfloor T/K \rfloor$ times, and results in $n$ residuals for each time step (Figures \ref{Fig2}.c). 

\begin{algorithm}[tb]
   \caption{BJ-based uncertainty estimation in RNNs}
   \label{alg:example}
\begin{algorithmic}[1] 
   \STATE {\bfseries Input:} Trained RNN parameters \mbox{\small $\hat{\boldsymbol{\theta}}(\boldsymbol{w})$}, data set \mbox{\small $\mathcal{D}_n$},
   \STATE $\,\,\,\,\,\,\,\,\,\,\,\,\,\,\,\,\,\,$ interval block length \mbox{\small $K$}, coverage level \mbox{\small $\alpha$},
   \STATE $\,\,\,\,\,\,\,\,\,\,\,\,\,\,\,\,\,\,$ test input sequence \mbox{\small $\boldsymbol{x}^*$}.
   \vspace{0.025in} 
   \STATE {\bfseries Output:} Confidence intervals \mbox{\small $\{\mathcal{C}_t(\mathcal{H}^*_t;\alpha)\}_t$}.\\ 
   \vspace{-0.05in}   
   \rule{8cm}{0.4pt}
   %\REPEAT
   \STATE \mbox{\small \texttt{// Resampling sequence and interval blocks}}
   %\vspace{0.015in}
   \FOR{$i=1$ {\bfseries to} $n$}
   \FOR{$j=1$ {\bfseries to} $\lfloor \frac{T}{K} \rfloor$}
   \vspace{0.035in}
   \STATE \mbox{\small $\boldsymbol{w}^\prime \gets \boldsymbol{w}^{\mbox{\tiny $-i$}}_{\mbox{\tiny $-j[K]$}}.$}
   \STATE \mbox{\small $\mathcal{I}_{\boldsymbol{w}}(\boldsymbol{w}^{\prime}) \gets$ \textsc{BlockwiseInfluence}$(\boldsymbol{w}, \boldsymbol{w}^{\prime}, \hat{\boldsymbol{\theta}}).$}
   \vspace{0.03in} 
   \STATE \mbox{\small $\hat{\boldsymbol{\theta}}(\boldsymbol{w}^{\prime}) \gets \hat{\boldsymbol{\theta}}(\boldsymbol{w}) + \mathcal{I}_{\boldsymbol{w}}(\boldsymbol{w}^{\prime}).$}
   \vspace{0.03in} 
   \FOR{$t=K\cdot j$ {\bfseries to} $K\cdot (j+1)$}
   \vspace{0.035in}
   \STATE \mbox{\small $r^{(i,j)}_{t, t^{\prime}} \gets \big|\,y^{(i)}_{t + t^{\prime}} - f_{t,t^{\prime}}(\boldsymbol{x}^{(i)}_{1:t}; \hat{\boldsymbol{\theta}}(\boldsymbol{w}^{\prime}))\,\big|,\, \forall t^{\prime}.$}
   \vspace{0.03in}
   \ENDFOR
   \ENDFOR
   \ENDFOR
   \vspace{0.025in}
   \STATE \mbox{\small \texttt{// Constructing confidence intervals}}
   \vspace{0.035in}
   \STATE Construct the prediction variability set \mbox{\small $\mathcal{V}_{t, t^{\prime}}(\boldsymbol{x}^{*},\gamma)$}.  
   \vspace{0.035in}   
   \STATE \mbox{\small $f^U_{t,t^{\prime}} \gets \widehat{Q}^+_{\alpha}\{\mathcal{V}_{t, t^{\prime}}(\boldsymbol{x}^{*};+1)\},\, \forall t, t^{\prime}.$}
   \vspace{0.035in}
   \STATE \mbox{\small $f^L_{t,t^{\prime}} \gets \widehat{Q}^-_{\alpha}\{\mathcal{V}_{t, t^{\prime}}(\boldsymbol{x}^{*};-1)\},\, \forall t, t^{\prime}.$}
   \vspace{0.035in}
   \STATE {\bf Return} \mbox{\small $\mathcal{C}_{t,t^{\prime}} = [f^L_{t,t^{\prime}}, f^U_{t,t^{\prime}}],\, \forall t, t^\prime.$}
   %\vspace{0.025in}
\end{algorithmic} 
\end{algorithm} 

\subsection{Constructing Confidence Intervals}
\label{Sec42} % Refer to Algorithm
The sequence of confidence intervals $\{\mathcal{C}_{t,t^{\prime}}\}_t$ is a {\it bivariate} stochastic process on $(\Omega, \mathcal{H}, \mathbb{P})$ that ``sandwiches'' the true output over time. Given the resampled RNN parameters $\{\hat{\boldsymbol{\theta}}(\boldsymbol{w}^{\mbox{\tiny $-i$}}_{\mbox{\tiny $-j[K]$}})\}_{i,j}$ obtained through the BJ procedure in Section \ref{Sec41}, we describe how to construct the confidence intervals $\{\mathcal{C}_{t,t^{\prime}}\}_t$ in this Section. Before proceeding, we first define some notation. For any set $\mathcal{V} = \{v_i\}^n_{i=1}$,~define~the $(1-\alpha)$ and $\alpha$ empirical quantiles $\widehat{Q}^+_{\alpha}\{\mathcal{V}\}$ and $\widehat{Q}^-_{\alpha}\{\mathcal{V}\}$ as
\begin{align}
\widehat{Q}^+_{\alpha}\{\mathcal{V}\} &\triangleq \mbox{the $\lceil (1-\alpha)(n+1) \rceil$-th smallest value in $\mathcal{V}$,} \nonumber\\
\widehat{Q}^-_{\alpha}\{\mathcal{V}\} &\triangleq \mbox{the $\lfloor \alpha(n+1) \rfloor$-th smallest value in $\mathcal{V}$.} \nonumber
\end{align}
For a test sequence \mbox{\small $\boldsymbol{x}^* \in \mathcal{X}^T$}, define \mbox{\small $\mathcal{V}_{t, t^{\prime}}(\boldsymbol{x}^{*},\gamma) = \{v^{(i)}_{t, t^{\prime}}\}_{i}$} as a {\it prediction variability} set, with \mbox{\small $\gamma \in \{-1,+1\}$}, where the elements of \mbox{\small $\mathcal{V}_{t, t^{\prime}}$} are constructed as \cite{barber2019predictive}
\begin{align}
v^{(i)}_{t, t^{\prime}}(\boldsymbol{x}^{*}_{1:t};\gamma) = \underbrace{f_{t,t^{\prime}}(\boldsymbol{x}^*_{1:t}; \hat{\boldsymbol{\theta}}(\boldsymbol{w}^{\mbox{\tiny $-i$}}_{\mbox{\tiny $-j[K]$}}))}_{\mbox{\small {\bf Path-dependent}}} \,\, + \underbrace{\gamma \cdot r^{(i,j)}_{t, t^{\prime}},}_{\mbox{\small {\bf Time-dependent}}} \nonumber
\end{align} 
with \mbox{\small $j$} satisfying \mbox{\small $K\cdot j \leq t \leq K\cdot (j+1)$} for a given \mbox{\small $t$}. Each element in \mbox{\small $\mathcal{V}_{t, t^{\prime}}$} comprises two component. The first component represents the {\it path-dependent} variability of the resampled RNN outputs on the new test sequence $\boldsymbol{x}^{*}$ (pink dots in Figure \ref{Fig2}.c) --- this reflects the uncertainty of the model about its prediction for this specific path of the stochastic processes in $(\Omega, \mathcal{H}, \mathbb{P})$, i.e., the variability will be larger if $\boldsymbol{x}^{*}$ is a ``rarely seen'' sequence in the training data \mbox{\small $\mathcal{D}_n$}. The second component is the set of LOBO residuals of the RNN model on the training data, which depend only on the time step $t$, and are fixed for any new test sequence (blue dots in Figure \ref{Fig2}.c). The residual terms reflect how well the RNN model generalizes to a new sequence at each time step $t$.~The lower limit of the confidence interval \mbox{\small $\mathcal{C}_{t,t^{\prime}}=[f^L_{t,t^{\prime}},f^U_{t,t^{\prime}}]$} is constructed as \mbox{\small $f^L_{t,t^{\prime}} = \widehat{Q}^-_{\alpha}\{\mathcal{V}_{t, t^{\prime}}(\boldsymbol{x}^{*};-1)\}$}, whereas the upper~limit~on~confidence is \mbox{\small $f^U_{t,t^{\prime}} = \widehat{Q}^+_{\alpha}\{\mathcal{V}_{t, t^{\prime}}(\boldsymbol{x}^{*};+1)\}$}. As we will show in Section \ref{Sec44}, constructing confidence intervals using the quantiles of the prediction variability set \mbox{\small $\mathcal{V}_{t, t^{\prime}}$} ensures that the coverage condition in (\ref{Eq4}) is satisfied.   

%The significance of the two types of data blocks vary from one application to another. In settings where data comprises many realizations of a stochastic process, e.g., medical time-series data for many patients \cite{shickel2017deep}, then resampling over sequence blocks can be sufficient. On the contrary, in applications where the RNN is trained on few long time-series, e.g., stock prices, then interval blocks are more significant. In the special case when $n=1$, BJ resampling reduces to the classical fixed-block scheme proposed by K\"unsch in \cite{kunsch1989jackknife,busing1999delete}. 

\newpage
\subsection{Blockwise Influence Functions}
\label{Sec43} 
Implementing exact BJ resampling requires re-training the RNN model \mbox{\small $n \cdot \lfloor T/K \rfloor$} times,~which~is~computationally~infeasible for most relevant applications that entail large data sets and long sequences. To avoid re-training the RNN, we use influence functions to estimate \mbox{\small $\hat{\boldsymbol{\theta}}(\boldsymbol{w}^{-i}_{\mbox{\tiny $-j[K]$}})$} without explicit model fitting --- this corresponds to an ``infinitesimal'' version of our jackknife scheme \cite{efron1992jackknife}.  

The {\it blockwise} influence function, denoted as \mbox{\small $\mathcal{I}_{\boldsymbol{w}}(\boldsymbol{w}^{\prime})$}, quantifies the change in RNN parameters when training using the weight vector $\boldsymbol{w}^{\prime}$ instead of $\boldsymbol{w}$ through a first-order Taylor approximation given by \cite{hampel2011robust}: 
\begin{align}
\hat{\boldsymbol{\theta}}(\boldsymbol{w}^{\prime}) \approx \hat{\boldsymbol{\theta}}(\boldsymbol{w}) + \mathcal{I}_{\boldsymbol{w}}(\boldsymbol{w}^{\prime}),
\label{Eq7} 
\end{align}
and can be computed as follows \cite{koh2019accuracy}:
\begin{align}
\mathcal{I}_{\boldsymbol{w}}(\boldsymbol{w}^{\prime}) = H^{-1}_{\boldsymbol{w}}\,\, g_{\boldsymbol{w}}(\boldsymbol{w}^{\prime}),
\label{Eq8} 
\end{align}
where $H_{\boldsymbol{w}} = \nabla^2_\theta\, L(\mathcal{D}_n; \hat{\boldsymbol{\theta}}(\boldsymbol{w}), \boldsymbol{w})$ is the Hessian of the loss function, and $g_{\boldsymbol{w}}(\boldsymbol{w}^\prime) = \nabla_\theta\, L(\mathcal{D}_n; \hat{\boldsymbol{\theta}}(\boldsymbol{w}), \boldsymbol{w}^\prime)$.~In~what~follows, we develop an algorithm for efficiently computing (\ref{Eq8}) (the function \mbox{\small \textsc{BlockwiseInfluence}} in Algorithm 1). 

{\bf Computing Blockwise Influence.} The bottleneck in computing the influence function in (\ref{Eq8}) is the inversion of the Hessian matrix $H_{\boldsymbol{w}}$; for $P$ RNN parameters in $\hat{\boldsymbol{\theta}}$, the complexity of computing $H_{\boldsymbol{w}}^{-1}$ is $\mathcal{O}(P^3)$. Moreover, because the RNN parameters are shared across time steps, the complexity of gradient computation using backpropagation through time (BPTT) is $\mathcal{O}(PT)$. This computational burden~can~be significant when performing interval block deletion on long sequences, since deleting the $j$-th interval block entails altering the RNN hidden states in time steps $t \geq K\cdot (j+1)$ and re-computing all gradients for those time steps.  

\begin{figure}[t]
  \centering
  \includegraphics[width=3.25in]{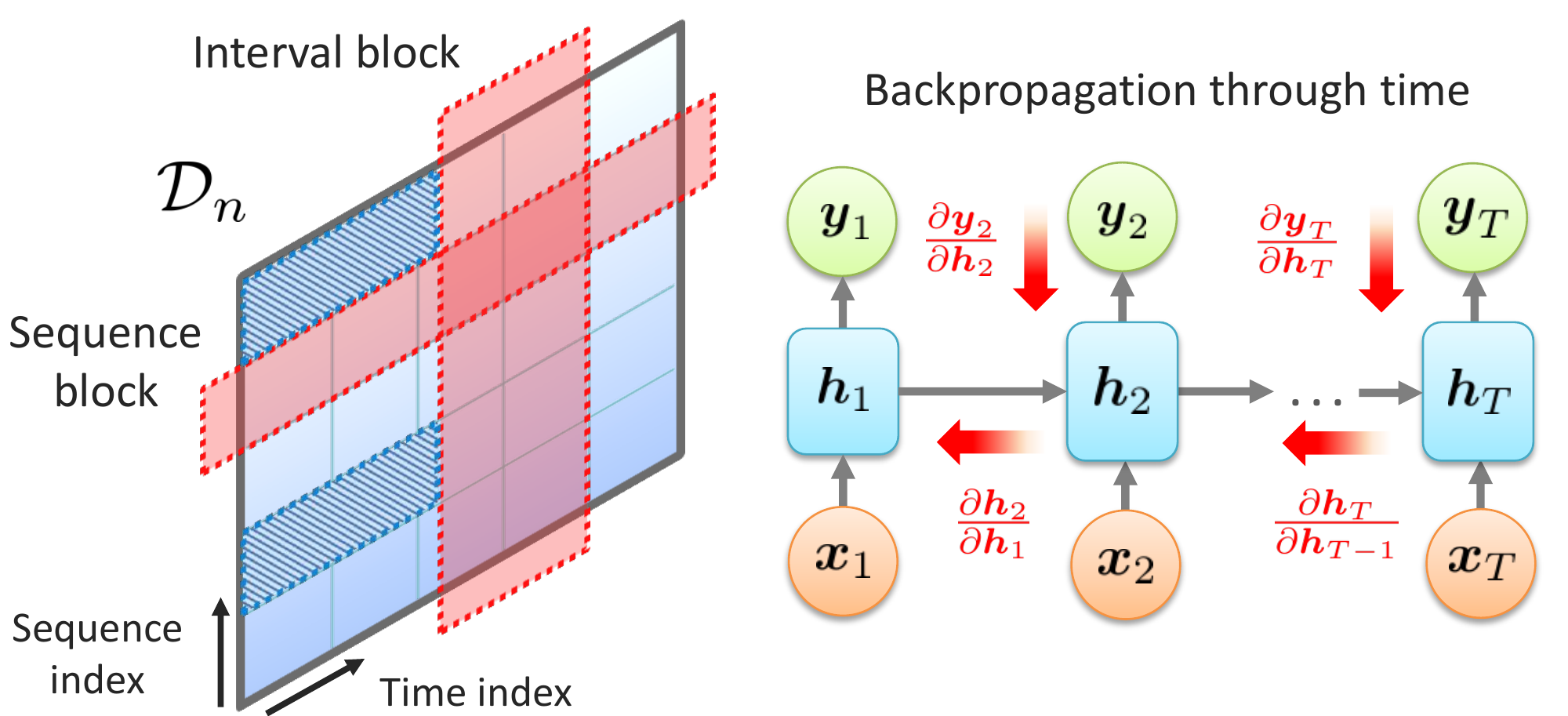}
  \caption{\footnotesize Illustration for blockwise batch sub-sampling in our HVP stochastic estimation algorithm. {\bf Left:} Red-shaded areas are the deleted sequence and interval blocks and blue-shaded areas are the sampled blocks for gradient computation. {\bf Right:} Gradient computation in BPTT --- due to parameter sharing, gradients at time $t$ depend on gradients at all preceding time steps $t^\prime <t$.} 
  \rule{\linewidth}{.75pt}   
  \label{Fig3} 
  \vspace{-7mm} 
\end{figure}

Following the approaches in \cite{pearlmutter1994fast, agarwal2016second,koh2017understanding}, we address the computational challenges above by developing a stochastic estimation approach with blockwise batch sampling to~estimate~the hessian-vector product (HVP) \mbox{\small $H^{-1}_{\boldsymbol{w}}\,\, g_{\boldsymbol{w}}(\boldsymbol{w}^{\prime})$}, without explicitly inverting or forming the Hessian.~The~pseudo-code~for \mbox{\small \textsc{BlockwiseInfluence}}, which computes $\mathcal{I}_{\boldsymbol{w}}(\boldsymbol{w}^{\mbox{\tiny $-i$}}_{\mbox{\tiny $-j[K]$}})$, is:

\mbox{\tiny $\blacksquare$}\,\, Randomly sample $v \leq n-1$ sub-sequences $m_1,\ldots,m_v$, each represented as a tuple \mbox{\small $\{\mathcal{S}^{(m_s)}_{1:Kj}, \mathcal{S}^{(m_s)}_{K(j+1):T}\}^v_{s=1}$},~with~the $j$-th interval block deleted, then from each tuple select \mbox{\small $\mathcal{S}^{(m_s)}_{1:Kj}$}. In doing so,~we~only~include temporally contiguous data in the sampled batch, hence retaining the RNN hidden states computed in training. (See Figure \ref{Fig3} for an illustration.)

%\mbox{\tiny $\blacksquare$}\,\, Evaluate the re-weighted gradients $g=g_{\boldsymbol{w}}(\boldsymbol{w}^{\mbox{\tiny $-i$}}_{\mbox{\tiny $-j[K]$}})$. 

\mbox{\tiny $\blacksquare$}\,\, For all $j$ and $s \in \{1,\ldots,v\}$, recursively compute 
\begin{align}
\widetilde{H}^{-1}_{s,\boldsymbol{w}}\,g = g + (\boldsymbol{I} - \nabla^2_\theta\, L(\mathcal{S}^{(m_s)}_{1:Kj}; \hat{\boldsymbol{\theta}}(\boldsymbol{w}), \boldsymbol{w}^\prime))\,\widetilde{H}^{-1}_{s-1,\boldsymbol{w}}\,g, \nonumber
\end{align}
where $g=g_{\boldsymbol{w}}(\boldsymbol{w}^{\mbox{\tiny $-i$}}_{\mbox{\tiny $-j[K]$}})$ is the loss gradient after $j$-th interval and $i$-th sequence deletion, $\widetilde{H}^{-1}_{s,\boldsymbol{w}} = \sum_{v=0}^{s}(\boldsymbol{I}-\widetilde{H}_{s,\boldsymbol{w}})^{v}$, and $\widetilde{H}$ is the stochastic estimate of the Hessian $H$. The recursive procedure is initialized with $\widetilde{H}^{-1}_{0,\boldsymbol{w}}\,g = g$, and our final estimate of the HVP is $\widetilde{H}^{-1}_{v,\boldsymbol{w}}\,g$. The recursion above follows from the power series expansion of a matrix inverse, which converges when $H_{\boldsymbol{w}} \succ 0$ and all its eigenvalues are less than 1. Since restricting our sampled sub-sequences to the interval $[1,Kj]$ enables reusing the loss gradients evaluated in training, the overall complexity of our procedure is $\mathcal{O}(nP+vP)$. Further discussion on the complexity and stability of the procedure above is given in Appendix B.  

\subsection{Theoretical Guarantees}
\label{Sec44} 
We conclude this Section with a Theorem that~characterizes the time-dependent coverage performance of the BJ-based confidence intervals constructed via Algorithm 1.

{\bf Theorem 1.} {\it Let $\beta_{T}(\tau)$ be the $\beta$-mixing coefficient of the stochastic processes drawn from $(\Omega,\mathcal{H},\mathbb{P})$, and let $T_0 \leq T$ be the time interval for which $\beta_{T}(T_0)=0$. For an interval block length of $K=T_0$, the time-dependent coverage achieved by the BJ procedure with exact re-training is}
\[\mathbb{P}(\varphi(y_{t+t^{\prime}}) \in \mathcal{C}_{t,t^{\prime}}(\mathcal{H}_t;\alpha)) \geq (1 - 2\alpha),\, \forall 1 \leq t \leq T,\]
{\it for~prediction~horizon~$t^\prime$,~and coverage $\alpha\in (0,1)$.} $\,\,\,\,\,\,\,\,\,\,\square$

Theorem 1 shows that if the interval block length $K$ matches the inherent correlation structure of the data, i.e., $K$ is equal to the ``mixing-time'' $T_0$, then confidence intervals generated by exact BJ resampling always achieve a time-dependent coverage of $(1-2\alpha)$ --- as we show in the next Section, BJ resampling usually achieves the target $(1-2\alpha)$ in practice \cite{barber2019predictive}. The extent to which the approximate (influence function-based) BJ is capable of achieving the target coverage depends on the quality of estimating the re-trained parameters, which in turn depends on various properties of the model's loss function (e.g., convexity and differentiability \cite{giordano2018swiss, koh2019accuracy}).  

\begin{figure*}[t]
  \centering
  \includegraphics[width=6.75in]{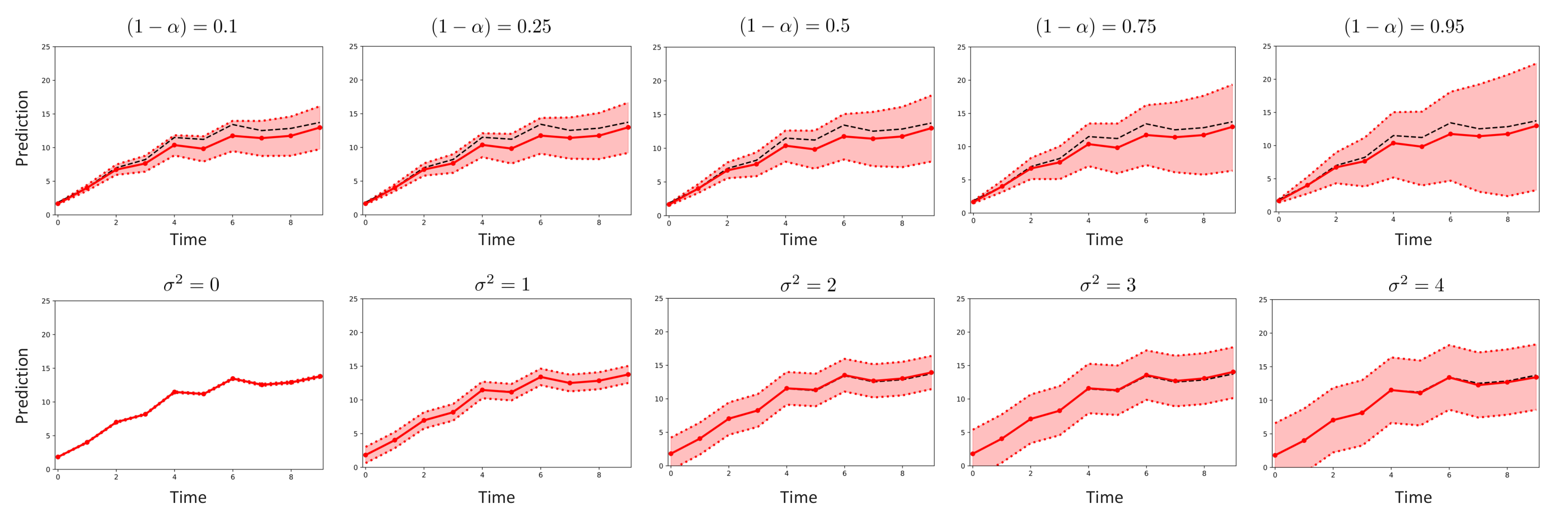}
  \caption{\footnotesize {\bf Sequential confidence intervals.} (a) {\it Time-dependent aleatoric uncertainty.} In the top panels,~we~display~the~sequential~confidence intervals $\{\mathcal{C}_{t,1}\}_t$ issued on top of a vanilla RNN model over $T=10$ time steps within a single sequence. Here,~the~model~is~trained~on data generated by the process in (\ref{eqSec51}) using the time-dependent noise profile in (\ref{eqSec52}). The red lines correspond to the RNN predictions~and~the dashed black lines correspond to the true labels $\{y_t\}_t$. We display the confidence intervals at different~coverage~levels:~as~we~can~see,~the confidence intervals get wider over time to cover the progressively increasing aleatoric noise. The confidence intervals get wider for higher coverage levels. (b) {\it Static aleatoric uncertainty.} In the bottom panels, we display the confidence intervals for the same test sequence in the top panels but with the static noise profile in (\ref{eqSec52}). Here, we trained 5 RNN models with varying levels of the static noise variance $\bar{\sigma}^2 \in \{0,1,2,3,4\}$. As we can see, the sequential confidence intervals are fixed over time, reflecting the uniform aleatoric noise across all time steps. The confidence intervals vanish when no noise is added ($\bar{\sigma}^2=0$) and gets increasingly wide as the noise variance increases.} 
  \rule{\linewidth}{.75pt}   
  \label{Fig_new_1} 
  \vspace{-7mm} 
\end{figure*} 

\newpage
\section{Experiments}
\label{Sec5}
In this Section, we evaluate our method~quantitatively~and qualitatively through synthetic simulations~(Section~\ref{Sec51}) and experiments on real-world data (Section \ref{Sec52}). Throughout this Section, we set $K=T$ in all experiments. 

\subsection{Synthetic Simulations}
\label{Sec51}
We consider a synthetic model~for~generating~sequential~data through the following autoregressive process:
\vspace{-0.1in}
\begin{align}
y_t = \sum_{k=0}^{t} a^k \cdot x_{k} + \epsilon_t,\, k \in \{1,\ldots,T\}, 
\label{eqSec51}
\end{align}
where $a=0.9$ and $\epsilon_k \sim \mathcal{N}(0, \sigma^2_k)$ is a~noise~process,~with~a variance $\sigma^2_t$ at time step $t$. Throughout our experiments, we consider the following noise profiles:
\begin{align}
\mbox{\it Static noise variance}:\,\,& \sigma^2_t = \bar{\sigma}^2, \nonumber \\ 
\mbox{\it Time-dependent noise variance}:\,\,& \sigma^2_t = t/10. 
\label{eqSec52}
\end{align}
The first noise profile induces a fixed level of aleatoric uncertainty across all time steps, whereas~the~second~noise~profile induces an increasing level of aleatoric uncertainty in the data as time progresses within the same sequence.

{\bf Baselines.} We applied our BJ procedure (in a post-hoc manner) to standard RNN trained to learn one-step-ahead forecasts ($T^{\prime}=1$). We compared uncertainty estimates obtained by Algorithm 1 with 2 baselines that cover the Bayesian and Frequentist uncertainty estimation methods discussed in Section \ref{Sec1}. These included~the~multi-horizon~quantile~recurrent forecaster (MQ-RNN) in \cite{wen2017multi} and the Monte Carlo dropout-based RNNs (DP-RNN) \cite{gal2016theoretically}. To ensure a fair comparison, we fix the hyperparameters of the underlying RNNs in all baselines. Details on hyperparameter tuning and uncertainty calibration in all baselines are provided in Appendix D. Unless otherwise stated, confidence intervals by all baselines were estimated for a target coverage of $90\%$ ($\alpha=0.1$).  

\begin{figure*}[t] 
  \centering
  \subfigure{% 
    \includegraphics[width=.31\textwidth]{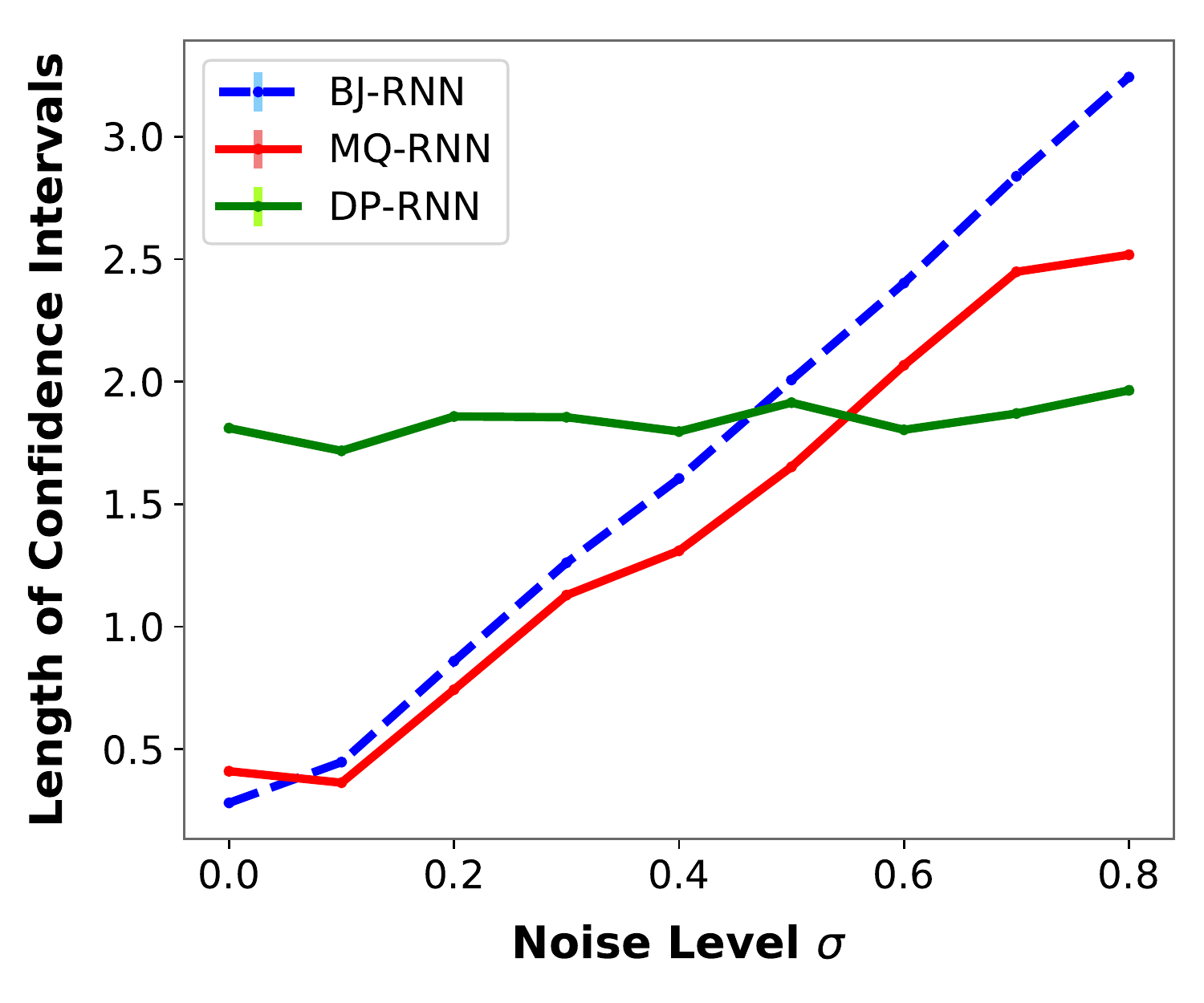} \label{fig:1} 
  } 
  \quad 
  \subfigure{% 
    \includegraphics[width=.31\textwidth]{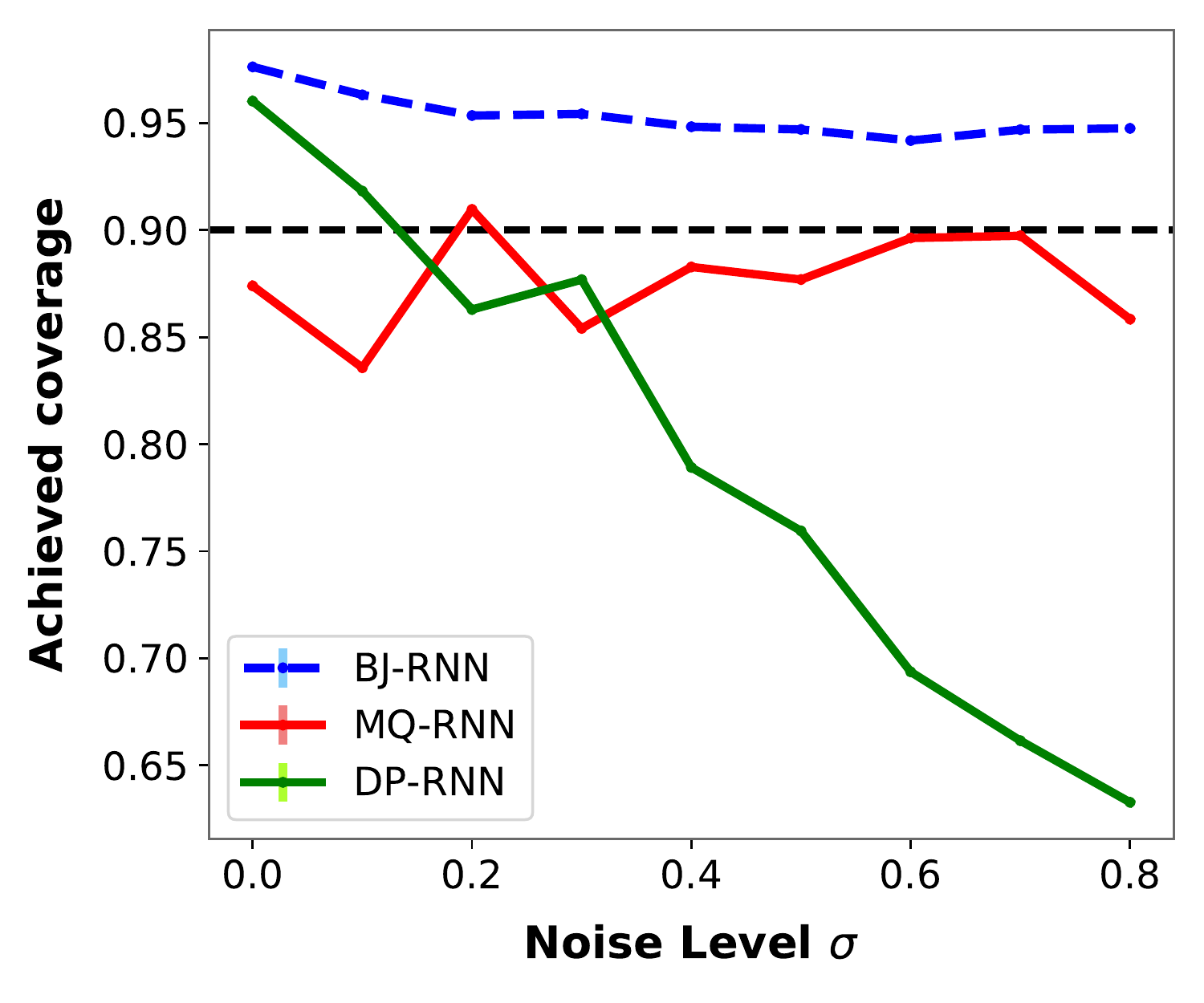} \label{fig:2} 
  } 
  \quad
  \subfigure{% 
    \includegraphics[width=.31\textwidth]{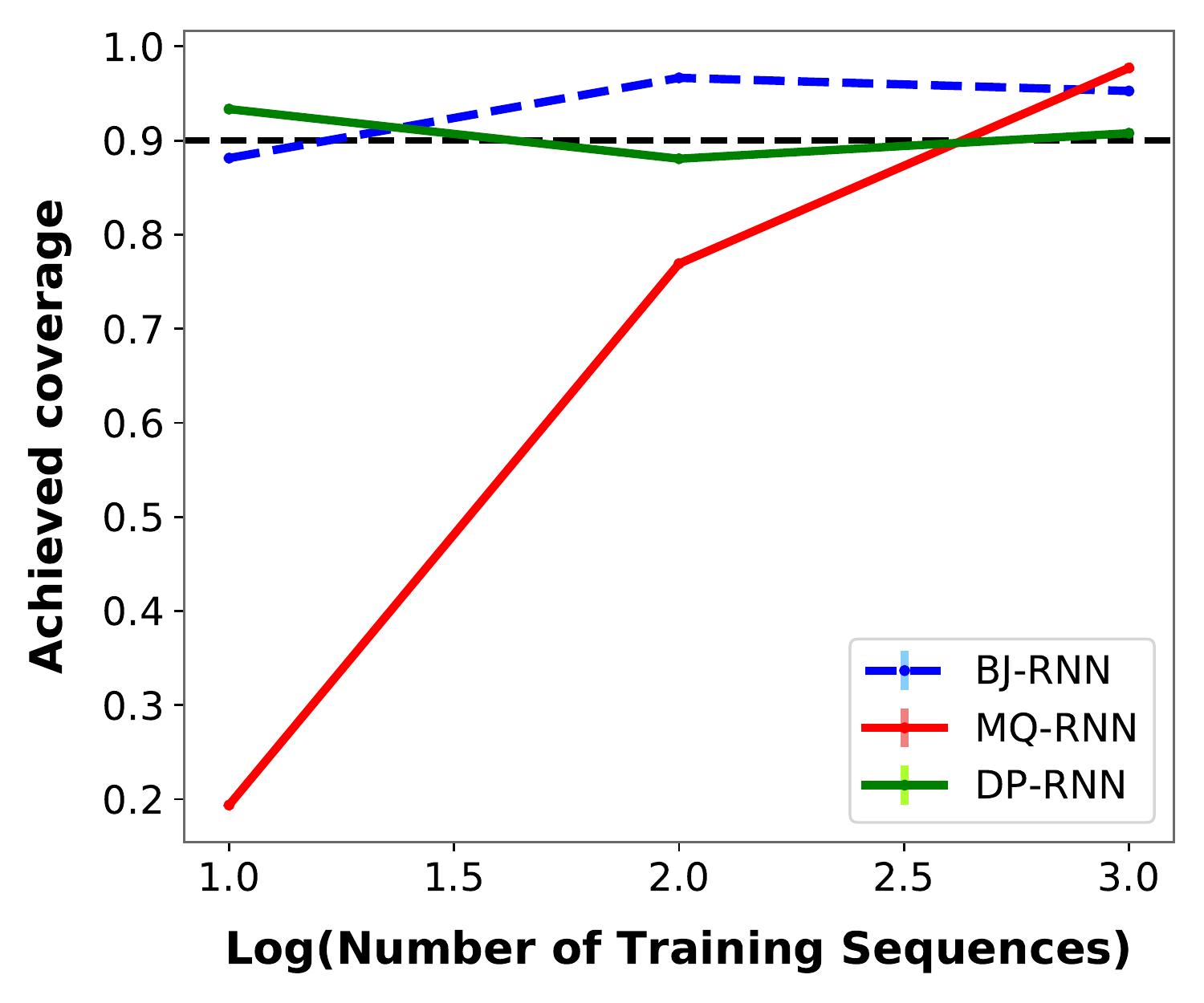} \label{fig:2} 
  } 
  \vspace{-6mm}
  \caption{\footnotesize {\bf Comparison between baselines at varying levels of aleatoric and epistemic uncertainty.} Black dashed lines are the target coverage rates. {\bf (Left)} Average length of confidence intervals issued by all baselines at different noise levels. {\bf (Middle)} Average coverage for all baselines at different noise levels. {\bf (Right)} Average coverage for all baselines for different numbers of training sequences.} 
\vspace{-1mm}  
  \rule{\linewidth}{.75pt}   
  \label{Fig_new_2} 
  \vspace{-8mm} 
\end{figure*}

{\bf Results.} We generated 1000 training sequences using the model in (\ref{eqSec51}) with $T=10$ time steps and trained a vanilla RNN model to predict the label $y_t$ on the basis of the input sequence $\boldsymbol{x}_{1:t}$. We conducted~this~experiment~using~the~time-dependent noise variance profile in (\ref{eqSec52}), in addition to multiple selections of the noise variance~for~the~static~noise~profile. In Figure \ref{Fig_new_1}, we display the sequential confidence intervals $\{\mathcal{C}_{t,1}\}_t$ issued by the BJ procedure under different simulation scenarios. As we can see, the sequential BJ confidence intervals reflect the ground-truth aleatoric uncertainty profiles dictated by the underlying~noise~processes.~That~is, the confidence intervals (1) are wider within a sequence as the noise variance increases over time for a time-dependent noise profile, (2) are of fixed width over time within a sequence for a static noise profile, and (3) have their width increase with the target coverage and the noise variance. 

In Figure \ref{Fig_new_2}, we examine the performance~of~all~baselines at varying levels of aleatoric and~epistemic~uncertainty.~Here we use the static noise profile in (\ref{eqSec52}). Aleatoric uncertainty is controlled by varying the level of noise $\bar{\sigma}^2$, whereas epistemic uncertainty is controlled by varying~the~number~of training sequences. As we can see, all~baselines~display an increasing length of confidence~intervals~with~increasing aleatoric uncertainty, except for DP-RNN which does not exhibit sensitivity to the increasing ground-truth uncertainty (Figure \ref{Fig_new_2}, left). Our model (BJ-RNN) achieves the target coverage for all levels of aleatoric uncertainty, whereas DP-RNN tends to under-cover when the noise levels are large (Figure \ref{Fig_new_2}, middle). With respect to epistemic uncertainty, we note that since MQ-RNN ``learns'' the prediction intervals, its ability to achieve the target coverage depends on the amount of training data --- MQ-RNN under-covers the true labels when the number of training sequences is small. On the contrary, since BJ-RNN is a post-hoc procedure, it achieves the target coverage irrespective of the size of training data (Figure \ref{Fig_new_2}, right).   
  
\subsection{Experiments on Real Data}
\label{Sec52}
How can uncertainty quantification help us conquer practical (sequential) decision-making problems? In this Section, we evaluate the utility of our procedure in guiding therapeutic decisions for critical care medicine. 

{\bf Data.} We conducted experiments on data from the Medical Information Mart for Intensive Care (MIMIC-III) \cite{johnson2016mimic} database, which comprises the electronic health records for critically-ill patients admitted to an intensive care unit (ICU). From MIMIC-III, we~extracted~sequential~data for patients on antibiotics, with trajectories up to $T=10$ days (time steps) --- this resulted in a data set of~5,833~sequences. For all patients, we extracted 5~variables~(lab~tests and vital signs) collected during hospitalization. More details on data processing can be found in Appendix D. 

{\bf Clinical setup.} Based on each patient's 5 features (in the sequence $\boldsymbol{x}$), critical care teams need to (repeatedly) decide whether or not to administer {\it antibiotics} for a given patient to regulate their white blood cell count (WBCC). WBCC levels are clinically categorized into: Leucopenic ($<4\times 10^{9}/L$), normal ($4-10 \times 10^{9}/L$), and Leucemoid ($>10 \times 10^{9}/L$) ranges \cite{waheed2003white}.~High~(Leucemoid)~WBCC~is associated with severe illness and poor outcomes \cite{shuman2012hematologic} --- antibiotics can~prevent~WBCC from hitting the Leucemoid range, but also entail the risk of plummeting WBCC to the Leucopenic range. Thus, decisions on antibiotic administration should balance risks againts benefits. 

\begin{figure}[t]
  \centering
  \includegraphics[width=3in]{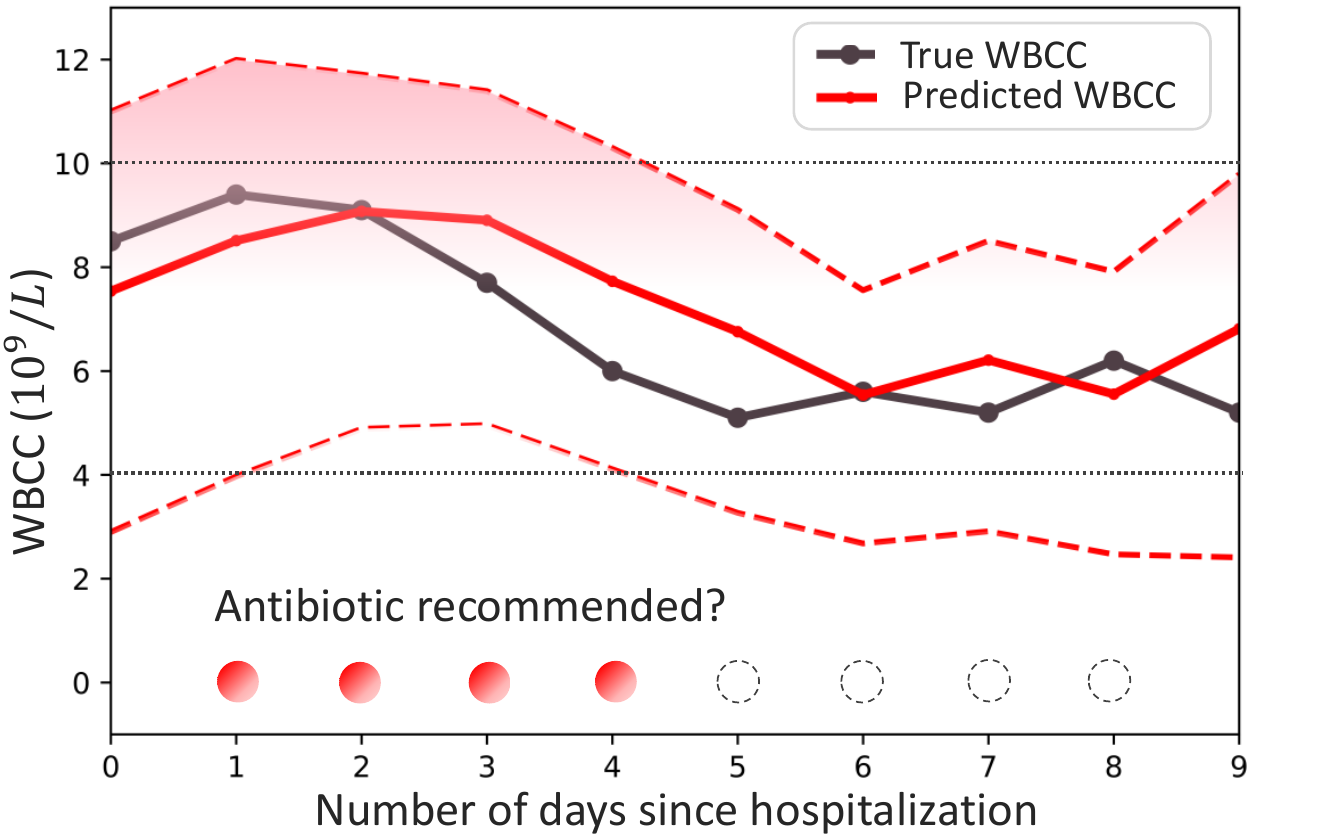}
  \caption{\footnotesize WBCC predictions issued by a standard RNN (\rxdash) for one patient over time, along with the BJ-based confidence intervals (dashed limits \rxdashed). An antibiotic is safely administered~(red~circle) on a given day if the upper confidence limit crosses the Leucopenic range, and the lower limit is above the Leucopenic range.} 
  \rule{\linewidth}{.75pt}   
  \label{Fig4} 
  \vspace{-5mm} 
\end{figure}

{\bf Evaluation.} We train various RNN-based~baselines~in~order to sequentially predict next-day's WBCC for every patient, and quantify the uncertainty in these predictions. (Figure \ref{Fig4} demonstrates the RNN predictions along with the~BJ-based confidence intervals for one patient in MIMIC-III.)~We~evaluate the uncertainty estimates by the~different~baselines~under study with respect to various performance metrics: the predictive accuracy of the underlying RNN model, the achieved coverage rate, the average length of the confidence interval and the accuracy of the uncertainty estimates in {\it auditing} model reliability. We use two metrics to~evaluate~the auditing accuracy of uncertainty estimates: the~{\it precision}~of predicting whether the true WBCC is in the Leucopenic range ($P_4$) and the precision of predicting whether WBCC is in the Leucemoid range ($P_{10}$). $P_4$ is evaluated using the lower confidence limit $f_L$ as the prediction and the indicator of whether WBCC $< 4\times 10^{9}/L$ as the label, whereas $P_{10}$ is evaluated with $f_U$ as the prediction and the indicator of whether WBCC $ > 10\times 10^{9}/L$ as the label. 

\begin{table}[t]
\centering
{\small
\begin{tabular}{@{}p{1.5cm}cccc@{}}
\toprule 
\multirow{2}{*}{\textbf{Methods}} & \multirow{2}{*}{\textbf{RMSE}} & \multirow{2}{*}{\textbf{Coverage}} &  \multirow{2}{*}{\textbf{CI length}} & \multicolumn{1}{c}{\textbf{Audit Acc.}} \\ [0.25ex]
  & &  & & \multicolumn{1}{c}{($P_4$, $P_{10}$)}   \\ \midrule 
 &   &     \\ [-2ex]
MQ-RNN     &  22.39   & 83.6$\%$ & 8.29 & (66$\%$, 42$\%$)  \\ [0.25ex]
DP-RNN     &  20.32   & 49.2$\%$ & 3.99 & (66$\%$, 50$\%$) \\ [0.25ex]
BJ-RNN & 20.24 & 94.4$\%$ & 13.18 & (67$\%$, 48$\%$) \\ [0.25ex]
 \bottomrule
\end{tabular}
\caption{\footnotesize Performance comparison between the different baselines. 
\label{Table2}}}
%\vspace{-2mm}
%\rule{\linewidth}{.75pt} 
\vspace{-5mm} 
\end{table}

{\bf Results.} In Table \ref{Table2}, we summarize the performance of all baselines. The first thing to notice is that since our BJ procedure is post-hoc, it was capable of providing accurate model audits without compromising the model's accuracy; the root mean square error (RMSE) of the vanilla RNNs were better than those of the baselines with built-in approaches to uncertainty quantification (MQ-RNN). In terms of the audit accuracy, BJ-based confidence intervals perform competitively with MQ-RNN --- which is specifically trained to optimize prediction intervals --- and the DP-RNN baseline. Moreover, because MQ-RNN is optimized to learn prediction intervals, it incurs a loss in predictive accuracy, whereas our model is applied in a post-hoc fashion without compromising the accuracy of the model's predictions. Notably, our procedure is the only baseline that is capable of achieving the desired coverage rate, whereas the two competing baselines undercover the true labels. The target coverage is achieved by our procedure without trivially exaggerating the lengths of the confidence intervals --- this is crucial for confidently informing clinicians on the true hematologic range for the patient at hand.

\section{Conclusion}
In this paper, we introduced one of the first approaches for estimating frequentist confidence intervals on the predictions of Recurrent neural networks (RNNs). Our method does not interfere with model training or compromise its accuracy, applies to any RNN architecture, and provides theoretical coverage guarantees on the estimated uncertainty intervals. The key idea behind our method is to re-sample ``blocks'' of temporally-correlated training data, and collect the predictions of the RNN re-trained on the remaining data. We used influence functions to approximate the re-trained RNN model parameters, thus eliminating the need for exhaustive re-training. Future extensions of our work would include more computationally efficient approaches for evaluating influence functions in RNN models with a large number of parameters, and analyzing the accuracy of blockwise influence functions as approximations of re-trained RNN model parameters. 

\section*{Acknowledgments}
The authors would like to thank the reviewers for their helpful comments. This work was supported by the US Office of Naval Research (ONR) and the National Science Foundation (NSF grants 1524417 and 1722516).

\bibliography{rnnjackknife_refs}

\begin{thebibliography}{62}
\providecommand{\natexlab}[1]{#1}
\providecommand{\url}[1]{\texttt{#1}}
\expandafter\ifx\csname urlstyle\endcsname\relax
  \providecommand{\doi}[1]{doi: #1}\else
  \providecommand{\doi}{doi: \begingroup \urlstyle{rm}\Url}\fi

\bibitem[Agarwal et~al.(2016)Agarwal, Bullins, and Hazan]{agarwal2016second}
Agarwal, N., Bullins, B., and Hazan, E.
\newblock Second-order stochastic optimization in linear time.
\newblock \emph{stat}, 1050:\penalty0 15, 2016.

\bibitem[Alaa \& Van Der~Schaar(2018)Alaa and Van Der~Schaar]{alaa2018hidden}
Alaa, A.~M. and Van Der~Schaar, M.
\newblock A hidden absorbing semi-markov model for informatively censored
  temporal data: Learning and inference.
\newblock \emph{The Journal of Machine Learning Research}, 19\penalty0
  (1):\penalty0 108--169, 2018.

\bibitem[Alaa \& van~der Schaar(2019)Alaa and van~der
  Schaar]{alaa2019attentive}
Alaa, A.~M. and van~der Schaar, M.
\newblock Attentive state-space modeling of disease progression.
\newblock In \emph{Advances in Neural Information Processing Systems}, pp.\
  11334--11344, 2019.

\bibitem[Auger \& Roy(2008)Auger and Roy]{auger2008expression}
Auger, A. and Roy, J.
\newblock Expression of uncertainty in linguistic data.
\newblock In \emph{2008 11th International Conference on Information Fusion},
  pp.\  1--8. IEEE, 2008.

\bibitem[Avati et~al.(2018)Avati, Duan, Jung, Shah, and Ng]{avati2018countdown}
Avati, A., Duan, T., Jung, K., Shah, N.~H., and Ng, A.
\newblock Countdown regression: sharp and calibrated survival predictions.
\newblock \emph{arXiv preprint arXiv:1806.08324}, 2018.

\bibitem[Bakker(2002)]{bakker2002reinforcement}
Bakker, B.
\newblock Reinforcement learning with long short-term memory.
\newblock In \emph{Advances in neural information processing systems}, pp.\
  1475--1482, 2002.

\bibitem[Bao et~al.(2017)Bao, Yue, and Rao]{bao2017deep}
Bao, W., Yue, J., and Rao, Y.
\newblock A deep learning framework for financial time series using stacked
  autoencoders and long-short term memory.
\newblock \emph{PloS one}, 12\penalty0 (7):\penalty0 e0180944, 2017.

\bibitem[Barber et~al.(2019{\natexlab{a}})Barber, Candes, Ramdas, and
  Tibshirani]{barber2019conformal}
Barber, R.~F., Candes, E.~J., Ramdas, A., and Tibshirani, R.~J.
\newblock Conformal prediction under covariate shift.
\newblock \emph{arXiv preprint arXiv:1904.06019}, 2019{\natexlab{a}}.

\bibitem[Barber et~al.(2019{\natexlab{b}})Barber, Candes, Ramdas, and
  Tibshirani]{barber2019predictive}
Barber, R.~F., Candes, E.~J., Ramdas, A., and Tibshirani, R.~J.
\newblock Predictive inference with the jackknife+.
\newblock \emph{arXiv preprint arXiv:1905.02928}, 2019{\natexlab{b}}.

\bibitem[Bayarri \& Berger(2004)Bayarri and Berger]{bayarri2004interplay}
Bayarri, M.~J. and Berger, J.~O.
\newblock The interplay of bayesian and frequentist analysis.
\newblock \emph{Statistical Science}, pp.\  58--80, 2004.

\bibitem[Chien \& Ku(2015)Chien and Ku]{chien2015bayesian}
Chien, J.-T. and Ku, Y.-C.
\newblock Bayesian recurrent neural network for language modeling.
\newblock \emph{IEEE transactions on neural networks and learning systems},
  27\penalty0 (2):\penalty0 361--374, 2015.

\bibitem[Dusenberry et~al.(2019)Dusenberry, Tran, Choi, Kemp, Nixon, Jerfel,
  Heller, and Dai]{dusenberry2019analyzing}
Dusenberry, M.~W., Tran, D., Choi, E., Kemp, J., Nixon, J., Jerfel, G., Heller,
  K., and Dai, A.~M.
\newblock Analyzing the role of model uncertainty for electronic health
  records.
\newblock \emph{arXiv preprint arXiv:1906.03842}, 2019.

\bibitem[Efron(1992)]{efron1992jackknife}
Efron, B.
\newblock Jackknife-after-bootstrap standard errors and influence functions.
\newblock \emph{Journal of the Royal Statistical Society: Series B
  (Methodological)}, 54\penalty0 (1):\penalty0 83--111, 1992.

\bibitem[Fedorov(2013)]{fedorov2013theory}
Fedorov, V.~V.
\newblock \emph{Theory of optimal experiments}.
\newblock Elsevier, 2013.

\bibitem[Fortunato et~al.(2017)Fortunato, Blundell, and
  Vinyals]{fortunato2017bayesian}
Fortunato, M., Blundell, C., and Vinyals, O.
\newblock Bayesian recurrent neural networks.
\newblock \emph{arXiv preprint arXiv:1704.02798}, 2017.

\bibitem[Gal(2016)]{gal2016uncertainty}
Gal, Y.
\newblock \emph{Uncertainty in deep learning}.
\newblock PhD thesis, PhD thesis, University of Cambridge, 2016.

\bibitem[Gal \& Ghahramani(2016{\natexlab{a}})Gal and
  Ghahramani]{gal2016dropout}
Gal, Y. and Ghahramani, Z.
\newblock Dropout as a bayesian approximation: Representing model uncertainty
  in deep learning.
\newblock In \emph{International Conference on Machine Learning (ICML)}, pp.\
  1050--1059, 2016{\natexlab{a}}.

\bibitem[Gal \& Ghahramani(2016{\natexlab{b}})Gal and
  Ghahramani]{gal2016theoretically}
Gal, Y. and Ghahramani, Z.
\newblock A theoretically grounded application of dropout in recurrent neural
  networks.
\newblock In \emph{Advances in neural information processing systems}, pp.\
  1019--1027, 2016{\natexlab{b}}.

\bibitem[Gasthaus et~al.(2019)Gasthaus, Benidis, Wang, Rangapuram, Salinas,
  Flunkert, and Januschowski]{gasthaus2019probabilistic}
Gasthaus, J., Benidis, K., Wang, Y., Rangapuram, S.~S., Salinas, D., Flunkert,
  V., and Januschowski, T.
\newblock Probabilistic forecasting with spline quantile function rnns.
\newblock In \emph{The 22nd International Conference on Artificial Intelligence
  and Statistics}, pp.\  1901--1910, 2019.

\bibitem[Gollier(2018)]{gollier2018economics}
Gollier, C.
\newblock \emph{The economics of risk and uncertainty}.
\newblock Edward Elgar Publishing Limited, 2018.

\bibitem[Graves et~al.(2006)Graves, Fern{\'a}ndez, Gomez, and
  Schmidhuber]{graves2006connectionist}
Graves, A., Fern{\'a}ndez, S., Gomez, F., and Schmidhuber, J.
\newblock Connectionist temporal classification: labelling unsegmented sequence
  data with recurrent neural networks.
\newblock 2006.

\bibitem[Hampel et~al.(2011)Hampel, Ronchetti, Rousseeuw, and
  Stahel]{hampel2011robust}
Hampel, F.~R., Ronchetti, E.~M., Rousseeuw, P.~J., and Stahel, W.~A.
\newblock \emph{Robust statistics: the approach based on influence functions},
  volume 196.
\newblock John Wiley \& Sons, 2011.

\bibitem[Hern{\'a}ndez-Lobato \& Adams(2015)Hern{\'a}ndez-Lobato and
  Adams]{hernandez2015probabilistic}
Hern{\'a}ndez-Lobato, J.~M. and Adams, R.
\newblock Probabilistic backpropagation for scalable learning of bayesian
  neural networks.
\newblock In \emph{International Conference on Machine Learning (ICML)}, pp.\
  1861--1869, 2015.

\bibitem[Hron et~al.(2017)Hron, Matthews, and Ghahramani]{hron2017variational}
Hron, J., Matthews, A. G. d.~G., and Ghahramani, Z.
\newblock Variational gaussian dropout is not bayesian.
\newblock \emph{arXiv preprint arXiv:1711.02989}, 2017.

\bibitem[Johnson et~al.(2016)Johnson, Pollard, Shen, Li-wei, Feng, Ghassemi,
  Moody, Szolovits, Celi, and Mark]{johnson2016mimic}
Johnson, A.~E., Pollard, T.~J., Shen, L., Li-wei, H.~L., Feng, M., Ghassemi,
  M., Moody, B., Szolovits, P., Celi, L.~A., and Mark, R.~G.
\newblock Mimic-iii, a freely accessible critical care database.
\newblock \emph{Scientific data}, 3:\penalty0 160035, 2016.

\bibitem[Kalchbrenner \& Blunsom(2013)Kalchbrenner and
  Blunsom]{kalchbrenner2013recurrent}
Kalchbrenner, N. and Blunsom, P.
\newblock Recurrent continuous translation models.
\newblock In \emph{Proceedings of the 2013 Conference on Empirical Methods in
  Natural Language Processing}, pp.\  1700--1709, 2013.

\bibitem[Koenker \& Hallock(2001)Koenker and Hallock]{koenker2001quantile}
Koenker, R. and Hallock, K.~F.
\newblock Quantile regression.
\newblock \emph{Journal of economic perspectives}, 15\penalty0 (4):\penalty0
  143--156, 2001.

\bibitem[Koh \& Liang(2017)Koh and Liang]{koh2017understanding}
Koh, P.~W. and Liang, P.
\newblock Understanding black-box predictions via influence functions.
\newblock In \emph{Proceedings of the 34th International Conference on Machine
  Learning-Volume 70}, pp.\  1885--1894. JMLR. org, 2017.

\bibitem[Koh et~al.(2019)Koh, Ang, Teo, and Liang]{koh2019accuracy}
Koh, P.~W., Ang, K.-S., Teo, H.~H., and Liang, P.
\newblock On the accuracy of influence functions for measuring group effects.
\newblock \emph{arXiv preprint arXiv:1905.13289}, 2019.

\bibitem[Krishnan et~al.(2017)Krishnan, Shalit, and
  Sontag]{krishnan2017structured}
Krishnan, R.~G., Shalit, U., and Sontag, D.
\newblock Structured inference networks for nonlinear state space models.
\newblock In \emph{Thirty-First AAAI Conference on Artificial Intelligence},
  2017.

\bibitem[Kunsch(1989)]{kunsch1989jackknife}
Kunsch, H.~R.
\newblock The jackknife and the bootstrap for general stationary observations.
\newblock \emph{The annals of Statistics}, pp.\  1217--1241, 1989.

\bibitem[Lakshminarayanan et~al.(2017)Lakshminarayanan, Pritzel, and
  Blundell]{lakshminarayanan2017simple}
Lakshminarayanan, B., Pritzel, A., and Blundell, C.
\newblock Simple and scalable predictive uncertainty estimation using deep
  ensembles.
\newblock In \emph{Advances in Neural Information Processing Systems
  (NeurIPS)}, pp.\  6402--6413, 2017.

\bibitem[Lawless \& Fredette(2005)Lawless and Fredette]{lawless2005frequentist}
Lawless, J. and Fredette, M.
\newblock Frequentist prediction intervals and predictive distributions.
\newblock \emph{Biometrika}, 92\penalty0 (3):\penalty0 529--542, 2005.

\bibitem[Lim et~al.(2018)Lim, Alaa, and Schaar]{lim2018forecasting}
Lim, B., Alaa, A., and Schaar, M. v.~d.
\newblock Forecasting treatment responses over time using recurrent marginal
  structural networks.
\newblock In \emph{Proceedings of the 32nd International Conference on Neural
  Information Processing Systems}, pp.\  7494--7504, 2018.

\bibitem[Luong et~al.(2016)Luong, Le, Sutskever, Vinyals, and
  Kaiser]{luong2015multi}
Luong, M.-T., Le, Q.~V., Sutskever, I., Vinyals, O., and Kaiser, L.
\newblock Multi-task sequence to sequence learning.
\newblock \emph{International Conference on Representation Learning (ICLR)},
  2016.

\bibitem[Maddox et~al.(2019)Maddox, Garipov, Izmailov, Vetrov, and
  Wilson]{maddox2019simple}
Maddox, W., Garipov, T., Izmailov, P., Vetrov, D., and Wilson, A.~G.
\newblock A simple baseline for bayesian uncertainty in deep learning.
\newblock \emph{arXiv preprint arXiv:1902.02476}, 2019.

\bibitem[Malinin \& Gales(2018)Malinin and Gales]{malinin2018predictive}
Malinin, A. and Gales, M.
\newblock Predictive uncertainty estimation via prior networks.
\newblock In \emph{Advances in Neural Information Processing Systems}, pp.\
  7047--7058, 2018.

\bibitem[Marra \& Wood(2012)Marra and Wood]{marra2012coverage}
Marra, G. and Wood, S.~N.
\newblock Coverage properties of confidence intervals for generalized additive
  model components.
\newblock \emph{Scandinavian Journal of Statistics}, 39\penalty0 (1):\penalty0
  53--74, 2012.

\bibitem[Mcdonald et~al.(2011)Mcdonald, Shalizi, and
  Schervish]{mcdonald2011estimating}
Mcdonald, D., Shalizi, C., and Schervish, M.
\newblock Estimating beta-mixing coefficients.
\newblock In \emph{Proceedings of the Fourteenth International Conference on
  Artificial Intelligence and Statistics}, pp.\  516--524, 2011.

\bibitem[Mentch \& Hooker(2016)Mentch and Hooker]{mentch2016quantifying}
Mentch, L. and Hooker, G.
\newblock Quantifying uncertainty in random forests via confidence intervals
  and hypothesis tests.
\newblock \emph{The Journal of Machine Learning Research (JMLR)}, 17\penalty0
  (1):\penalty0 841--881, 2016.

\bibitem[Miller(1974)]{miller1974jackknife}
Miller, R.~G.
\newblock The jackknife-a review.
\newblock \emph{Biometrika}, 61\penalty0 (1):\penalty0 1--15, 1974.

\bibitem[Mirikitani \& Nikolaev(2009)Mirikitani and
  Nikolaev]{mirikitani2009recursive}
Mirikitani, D.~T. and Nikolaev, N.
\newblock Recursive bayesian recurrent neural networks for time-series
  modeling.
\newblock \emph{IEEE Transactions on Neural Networks}, 21\penalty0
  (2):\penalty0 262--274, 2009.

\bibitem[Osband(2016)]{osband2016risk}
Osband, I.
\newblock Risk versus uncertainty in deep learning: Bayes, bootstrap and the
  dangers of dropout.
\newblock In \emph{NIPS Workshop on Bayesian Deep Learning}, 2016.

\bibitem[Ovadia et~al.(2019)Ovadia, Fertig, Ren, Nado, Sculley, Nowozin,
  Dillon, Lakshminarayanan, and Snoek]{ovadia2019can}
Ovadia, Y., Fertig, E., Ren, J., Nado, Z., Sculley, D., Nowozin, S., Dillon,
  J.~V., Lakshminarayanan, B., and Snoek, J.
\newblock Can you trust your model's uncertainty? evaluating predictive
  uncertainty under dataset shift.
\newblock \emph{arXiv preprint arXiv:1906.02530}, 2019.

\bibitem[Pearlmutter(1994)]{pearlmutter1994fast}
Pearlmutter, B.~A.
\newblock Fast exact multiplication by the hessian.
\newblock \emph{Neural computation}, 6\penalty0 (1):\penalty0 147--160, 1994.

\bibitem[Quenouille(1956)]{quenouille1956notes}
Quenouille, M.~H.
\newblock Notes on bias in estimation.
\newblock \emph{Biometrika}, 43\penalty0 (3/4):\penalty0 353--360, 1956.

\bibitem[Rangapuram et~al.(2018)Rangapuram, Seeger, Gasthaus, Stella, Wang, and
  Januschowski]{rangapuram2018deep}
Rangapuram, S.~S., Seeger, M.~W., Gasthaus, J., Stella, L., Wang, Y., and
  Januschowski, T.
\newblock Deep state space models for time series forecasting.
\newblock In \emph{Advances in Neural Information Processing Systems}, pp.\
  7785--7794, 2018.

\bibitem[Ritter et~al.(2018)Ritter, Botev, and Barber]{ritter2018scalable}
Ritter, H., Botev, A., and Barber, D.
\newblock A scalable laplace approximation for neural networks.
\newblock 2018.

\bibitem[Selvin et~al.(2017)Selvin, Vinayakumar, Gopalakrishnan, Menon, and
  Soman]{selvin2017stock}
Selvin, S., Vinayakumar, R., Gopalakrishnan, E., Menon, V.~K., and Soman, K.
\newblock Stock price prediction using lstm, rnn and cnn-sliding window model.
\newblock In \emph{2017 International Conference on Advances in Computing,
  Communications and Informatics (ICACCI)}, pp.\  1643--1647. IEEE, 2017.

\bibitem[Shickel et~al.(2017)Shickel, Tighe, Bihorac, and
  Rashidi]{shickel2017deep}
Shickel, B., Tighe, P.~J., Bihorac, A., and Rashidi, P.
\newblock Deep ehr: a survey of recent advances in deep learning techniques for
  electronic health record (ehr) analysis.
\newblock \emph{IEEE journal of biomedical and health informatics}, 22\penalty0
  (5):\penalty0 1589--1604, 2017.

\bibitem[Shuman et~al.(2012)Shuman, Demler, Trigoboff, and
  Opler]{shuman2012hematologic}
Shuman, M., Demler, T.~L., Trigoboff, E., and Opler, L.~A.
\newblock Hematologic impact of antibiotic administration on patients taking
  clozapine.
\newblock \emph{Innovations in clinical neuroscience}, 9\penalty0
  (11-12):\penalty0 18, 2012.

\bibitem[Sundermeyer et~al.(2012)Sundermeyer, Schl{\"u}ter, and
  Ney]{sundermeyer2012lstm}
Sundermeyer, M., Schl{\"u}ter, R., and Ney, H.
\newblock Lstm neural networks for language modeling.
\newblock In \emph{Thirteenth annual conference of the international speech
  communication association}, 2012.

\bibitem[Sutskever et~al.(2014)Sutskever, Vinyals, and
  Le]{sutskever2014sequence}
Sutskever, I., Vinyals, O., and Le, Q.
\newblock Sequence to sequence learning with neural networks.
\newblock \emph{Advances in NIPS}, 2014.

\bibitem[Taylor(2000)]{taylor2000quantile}
Taylor, J.~W.
\newblock A quantile regression neural network approach to estimating the
  conditional density of multiperiod returns.
\newblock \emph{Journal of Forecasting}, 19\penalty0 (4):\penalty0 299--311,
  2000.

\bibitem[Tukey(1958)]{tukey1958bias}
Tukey, J.
\newblock Bias and confidence in not quite large samples.
\newblock \emph{Ann. Math. Statist.}, 29:\penalty0 614, 1958.

\bibitem[Wager \& Athey(2018)Wager and Athey]{wager2018estimation}
Wager, S. and Athey, S.
\newblock Estimation and inference of heterogeneous treatment effects using
  random forests.
\newblock \emph{Journal of the American Statistical Association}, 113\penalty0
  (523):\penalty0 1228--1242, 2018.

\bibitem[Wager et~al.(2014)Wager, Hastie, and Efron]{wager2014confidence}
Wager, S., Hastie, T., and Efron, B.
\newblock Confidence intervals for random forests: The jackknife and the
  infinitesimal jackknife.
\newblock \emph{The Journal of Machine Learning Research (JMLR)}, 15\penalty0
  (1):\penalty0 1625--1651, 2014.

\bibitem[Waheed et~al.(2003)Waheed, Williams, Brett, Baldock, and
  Soni]{waheed2003white}
Waheed, U., Williams, P., Brett, S., Baldock, G., and Soni, N.
\newblock White cell count and intensive care unit outcome.
\newblock \emph{Anaesthesia}, 58\penalty0 (2):\penalty0 180--182, 2003.

\bibitem[Wang et~al.(2018)Wang, Zhang, He, and Zha]{wang2018supervised}
Wang, L., Zhang, W., He, X., and Zha, H.
\newblock Supervised reinforcement learning with recurrent neural network for
  dynamic treatment recommendation.
\newblock In \emph{Proceedings of the 24th ACM SIGKDD International Conference
  on Knowledge Discovery \& Data Mining}, pp.\  2447--2456. ACM, 2018.

\bibitem[Welling \& Teh(2011)Welling and Teh]{welling2011bayesian}
Welling, M. and Teh, Y.~W.
\newblock Bayesian learning via stochastic gradient langevin dynamics.
\newblock In \emph{Proceedings of the 28th international conference on machine
  learning (ICML-11)}, pp.\  681--688, 2011.

\bibitem[Wen et~al.(2017)Wen, Torkkola, Narayanaswamy, and
  Madeka]{wen2017multi}
Wen, R., Torkkola, K., Narayanaswamy, B., and Madeka, D.
\newblock A multi-horizon quantile recurrent forecaster.
\newblock \emph{arXiv preprint arXiv:1711.11053}, 2017.

\bibitem[Zhu \& Laptev(2017)Zhu and Laptev]{zhu2017deep}
Zhu, L. and Laptev, N.
\newblock Deep and confident prediction for time series at uber.
\newblock In \emph{2017 IEEE International Conference on Data Mining Workshops
  (ICDMW)}, pp.\  103--110. IEEE, 2017.

\end{thebibliography}
\bibliographystyle{icml2020}

\onecolumn

\appendix

\setcounter{figure}{0}
\setcounter{table}{0}
\setcounter{equation}{0}

\section*{Appendix}
\subsection*{Appendix A: Literature Survey}
Driven by the desire to suppress the ``overconfidence'' exhibited by poorly calibrated --- but highly discriminative --- deep learning models, various methods for uncertainty estimation in standard feed-forward neural networks have been recently proposed \cite{hernandez2015probabilistic,gal2016dropout,lakshminarayanan2017simple,malinin2018predictive}. However, the literature on uncertainty quantification in RNNs is rather scarce. In what follows, we provide a detailed overview of uncertainty quantification methods in RNNs, other methods that have been developed for feed-forward networks, background on jackknife resampling, and the connections between our work and modern applications of influence functions.

{\it Uncertainty quantification in RNNs}

We identified the following three strands of literature relating to uncertainty quantification in RNNs.

{\bf Bayesian RNNs.} The most prevalent approaches for RNN uncertainty estimation hinge on Bayesian modeling \cite{fortunato2017bayesian, chien2015bayesian, zhu2017deep,mirikitani2009recursive}. By specifying a prior over the RNN parameters, these approaches estimate a model's uncertainty via its posterior credible intervals. However, exact inference in Bayesian deep learning is generally intractable or computationally infeasible. Hence, practical Bayesian RNNs rely on dropout-based approximate inference \cite{gal2016dropout, gal2016theoretically}, which is often hard to calibrate, and is generally ill-posed as its posterior distributions do not concentrate asymptotically \cite{osband2016risk,hron2017variational}. Moreover, Bayesian methods require significant modifications to model architecture and training which limits their versatility. Our paper addresses these issues by developing a frequentist post-hoc alternative to Bayesian methods.

{\bf Quantile RNNs.} Similar to quantile regression \cite{koenker2001quantile}, quantile RNNs (Q-RNNs) learn prediction intervals by explicitly modeling the cumulative distribution function of the prediction targets \cite{taylor2000quantile, wen2017multi,gasthaus2019probabilistic}. The key difference between these approaches and ours is that Q-RNNs learn uncertainty jointly with the main prediction task, which as we have shown in Section \ref{Sec5}, can compromise their predictive accuracy. Moreover, Q-RNNs are bespoke models that must be retailored for each new prediction task, whereas our approach can be applied in a post-hoc manner to any RNN architecture. 

{\bf Probabilistic RNNs.} This class of models capture the structural uncertainty within the stochastic dynamics of sequential data. Existing variants of probabilistic RNNs are mainly based on deep state-space models \cite{krishnan2017structured, rangapuram2018deep,alaa2019attentive}. Probabilistic models are bespoke to the application at hand, and may not be appropriate for some application domains. Moreover, most probabilistic models entail restrictive assumptions on sequence dynamics, such as state Markovianity \cite{alaa2018hidden}.   

{\it Uncertainty quantification in feed-forward networks}

Bayesian models are the dominant tools for uncertainty quantification in feed-forward neural networks \cite{welling2011bayesian,hernandez2015probabilistic, ritter2018scalable, maddox2019simple}. Post-hoc application of Bayesian methods is not possible since Bayesian models require major modifications to the underlying predictive models (by specifying priors over model parameters). Moreover, Bayesian credible intervals do not guarantee frequentist coverage, and more crucially, the achieved coverage can be very sensitive to hyper-parameter tuning \cite{bayarri2004interplay}. Finally, a major limitation to exact Bayesian inference is that they are computationally prohibitive, and their alternative approximations --- e.g., \cite{gal2016dropout} --- may induced posteriors that do not concentrate asymptotically \cite{osband2016risk, hron2017variational}. 

Deep ensembles \cite{lakshminarayanan2017simple} are regarded as the most competitive alternative non-Bayesian benchmark for uncertainty estimation \cite{ovadia2019can}. These methods repeatedly re-trains the model on sub-samples of the data (using adversarial training), and then estimates uncertainty through the variance of the aggregate predictions. However, while applying deep ensemble methods may be feasible in feed-forward neural networks, creating large ensembles of RNN models is computationally infeasible, especially for long time series. 

{\it Jackknife resampling}

Jackknife resampling, originally developed by Quenouille in \cite{quenouille1956notes} and refined by Tukey in \cite{tukey1958bias}, is a general methodology for estimating sampling distributions. Our work builds on this literature by applying the jackknife to RNN predictions in order to estimate its predictive uncertainty through the sampling variance. The key difference between our setup and that of the standard jackknife is that RNNs do not operate on i.i.d data --- they operate on temporally correlated sequences. In \cite{kunsch1989jackknife}, K\"unsch proposed a variant of the jackknife for general stationary observations rather than i.i.d data. As we have shown in Section \ref{Sec5}, the sampling procedure therein is a special case of ours when the RNN is trained with a single data sequence (i.e., a single realization of the underlying stochastic process).

The (infinitesimal) jackknife method was previously used for quantifying the predictive uncertainty in random forests \cite{wager2014confidence,mentch2016quantifying,wager2018estimation}. In these works, however, the developed jackknife estimators are bespoke to bagging predictors, and cannot be straightforwardly extended to deep neural networks. More recently, general-purpose jackknife estimators were developed in \cite{barber2019predictive}, where two exhaustive leave-one-out procedures: the {\it jackknife+} and the {\it jackknife-minmax} where shown to have assumption-free worst-case coverage guarantees of $(1-2\alpha)$ and $(1-\alpha)$, respectively. However, our work is the first to develop such methods for time-series data. 

Previous application of influence functions has been limited to model interpretability by learning sample importance \cite{koh2017understanding, koh2019accuracy}. Our work uses the same machinery of influence function computation via Hessian-vector products, but for the purpose of estimating leave-one-out model parameters.  

\subsection*{Appendix B: Stochastic Estimation of HVPs via Blockwise Batch Sampling}

The bottleneck in computing the influence function in (\ref{Eq8}) is the inversion of the Hessian matrix $H_{\boldsymbol{w}}$; for $P$ RNN parameters in $\hat{\boldsymbol{\theta}}$, the complexity of computing $H_{\boldsymbol{w}}^{-1}$ is $\mathcal{O}(P^3)$. Moreover, because the RNN parameters are shared across time steps, the complexity of gradient computation using backpropagation through time (BPTT) is $\mathcal{O}(PT)$. This computational burden can be significant when performing interval block deletion on long sequences, since deleting the $j$-th interval block entails altering the RNN hidden states in time steps $t \geq K\cdot (j+1)$ and re-computing all gradients for those time steps.  

We start by the following well-known fact about the inverse of a matrix $H$ with $\|H\| \leq 1$ and $H \succ 0$. The inverse of such matrix can be evaluated through the following power series expansion:  
\begin{align}
H^{-1} = \sum^\infty_{i=0} (I-H)^i.
\end{align}

Following the approaches in \cite{pearlmutter1994fast, agarwal2016second,koh2017understanding}, we address the computational challenges above by developing a stochastic estimation approach with blockwise batch sampling to~estimate~the hessian-vector product (HVP) \mbox{\small $H^{-1}_{\boldsymbol{w}}\,\, g_{\boldsymbol{w}}(\boldsymbol{w}^{\prime})$}, without explicitly inverting or forming the Hessian.~The~pseudo-code~for \mbox{\small \textsc{BlockwiseInfluence}}, which computes $\mathcal{I}_{\boldsymbol{w}}(\boldsymbol{w}^{\mbox{\tiny $-i$}}_{\mbox{\tiny $-j[K]$}})$, is:

\mbox{\tiny $\blacksquare$}\,\, Randomly sample $v \leq n-1$ sub-sequences $m_1,\ldots,m_v$, each represented as a tuple \mbox{\small $\{\mathcal{S}^{(m_s)}_{1:Kj}, \mathcal{S}^{(m_s)}_{K(j+1):T}\}^v_{s=1}$},~with~the $j$-th interval block deleted, then from each tuple select \mbox{\small $\mathcal{S}^{(m_s)}_{1:Kj}$}. In doing so,~we~only~include temporally contiguous data in the sampled batch, hence retaining the RNN hidden states computed in training. (See Figure \ref{Fig3} for an illustration.)

\mbox{\tiny $\blacksquare$}\,\, For all $j$ and $s \in \{1,\ldots,v\}$, recursively compute 
\begin{align}
\widetilde{H}^{-1}_{s,\boldsymbol{w}}\,g = g + (I - \nabla^2_\theta\, L(\mathcal{S}^{(m_s)}_{1:Kj}; \hat{\boldsymbol{\theta}}(\boldsymbol{w}), \boldsymbol{w}^\prime))\,\widetilde{H}^{-1}_{s-1,\boldsymbol{w}}\,g, \nonumber
\end{align}
where $g=g_{\boldsymbol{w}}(\boldsymbol{w}^{\mbox{\tiny $-i$}}_{\mbox{\tiny $-j[K]$}})$ is the loss gradient after $j$-th interval and $i$-th sequence deletion, $\widetilde{H}^{-1}_{s,\boldsymbol{w}} = \sum_{v=0}^{s}(I-\widetilde{H}_{s,\boldsymbol{w}})^{v}$, and $\widetilde{H}$ is the stochastic estimate of the Hessian $H$. The recursive procedure is initialized with $\widetilde{H}^{-1}_{0,\boldsymbol{w}}\,g = g$, and our final estimate of the HVP is given by $\widetilde{H}^{-1}_{v,\boldsymbol{w}}\,g$, which converges to the true inverse for $v \to \infty$. 

Since restricting our sampled sub-sequences to the interval $[1,Kj]$ enables reusing the loss gradients evaluated in training, the overall complexity of our procedure is $\mathcal{O}(nP+vP)$, this can be broken down as follows. The complexity of multiplying $\nabla^2_\theta\, L(\mathcal{S}^{(m_s)}_{1:Kj}; \hat{\boldsymbol{\theta}}(\boldsymbol{w}), \boldsymbol{w}^\prime))$ with $\widetilde{H}^{-1}_{s-1,\boldsymbol{w}}\,g$ is $\mathcal{O}(P)$ per iteration which renders the overall complexity of computing this Hessian-product term as $\mathcal{O}(vP)$. The complexity of computing the initial gradient terms is $\mathcal{O}(nP)$. 

The procedure above converges only if the norm of the Hessian satisfies $\|H\| \leq 1$. Thus, if the largest eigenvalue $\sigma_{max}$ of $H$ is greater than 1, we normalize the loss to keep the largest eigenvalue less than 1. Moreover, we add a dampening term $\lambda$ to the Hessian in order to keep it invertible (positive definite). This results in the following recursion:
\begin{align}
\widetilde{H}^{-1}_{s,\boldsymbol{w}}\,g = g + (1-\lambda)\,\widetilde{H}^{-1}_{s-1,\boldsymbol{w}}\,g - \frac{1}{\sigma_{max}} \nabla^2_\theta\, L(\mathcal{S}^{(m_s)}_{1:Kj}; \hat{\boldsymbol{\theta}}(\boldsymbol{w}), \boldsymbol{w}^\prime)\,\widetilde{H}^{-1}_{s-1,\boldsymbol{w}}\,g, \nonumber
\end{align}
which converges with the appropriate selection of $\sigma_{max}$ and $\lambda$.

\newpage
\subsection*{Appendix C: Proof of Theorem 1}

Given a mixing-time of $T_0$, we can segment the data set $\mathcal{D}_n = \{(\boldsymbol{x}^{(i)}, y^{(i)})\}^n_{i=1}$ into evenly spaced segments of data 
\[\widetilde{\mathcal{D}}_n = \{(\boldsymbol{x}_{j\cdot T_0:j\cdot T_0+T_0}^{(i)}, y_{j\cdot T_0:j\cdot T_0+T_0}^{(i)})\}^n_{i=1},\]
which we can re-write as a new data set with $\tilde{n} = n\cdot T/T_0$ samples each with $d\cdot T_0$ dimensional features as follows:
\[\widetilde{\mathcal{D}}_n = \{(\tilde{\boldsymbol{x}}^{(j)}, \tilde{y}^{(j)})\}^{\tilde{n}}_{j=1},\]
where all the data points $\widetilde{\mathcal{D}}_n$ are independent. By treating each RNN prediction at time step $t$ as a new regression problem, it follows that the exact BJ procedure achieves coverage by directly applying Theorem 1 in \cite{barber2019conformal} to the modified data set $\widetilde{\mathcal{D}}_n$.

\subsection*{Appendix D: Experiments}

{\it Experimental Setup}

\begin{figure}[t]
  \centering
  \includegraphics[width=4.25in]{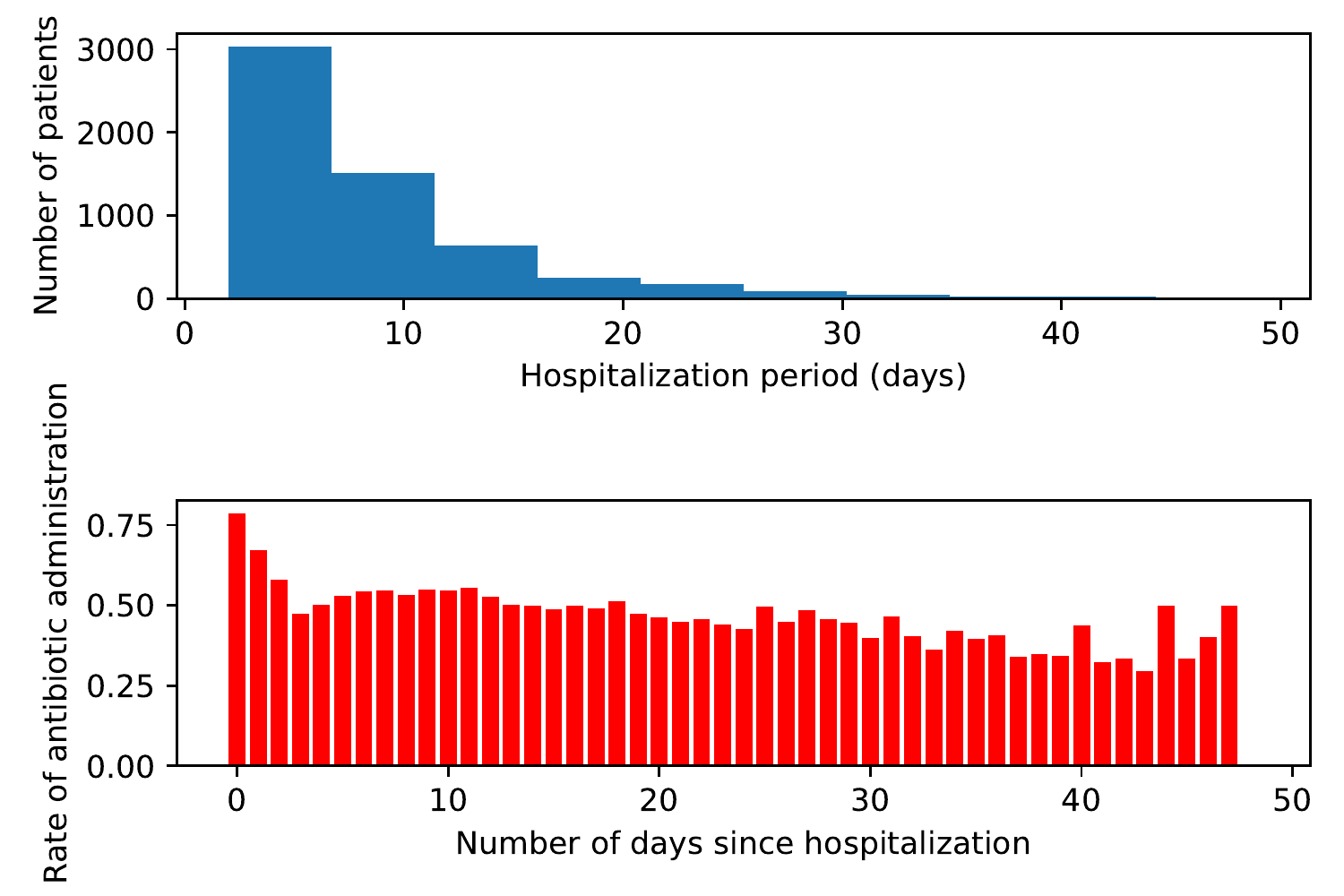}
  %\captionsetup{labelformat=empty, textformat=empty, labelsep=none}
  \caption{\footnotesize Statistics of hospitalization and treatment data.} 
  %\vspace{-8mm} 
  \label{FigD1}
\end{figure}

{\bf Data.} We used data from the Medical Information Mart for Intensive Care (MIMIC-III) \cite{johnson2016mimic} database, which comprises the electronic health records for critically-ill patients admitted to an intensive care unit (ICU). From MIMIC-III, we~extracted~sequential~data for patients on antibiotics, with trajectories up to $T=10$ time steps --- this resulted in a data set with $n=5,833$ sequences. For each patient, we extracted 30 variables (lab tests and vital signs) listed in Table \ref{FigD1}, based on which predictions are issued. 

\begin{figure}[h]
  \centering
  \includegraphics[width=3.25in]{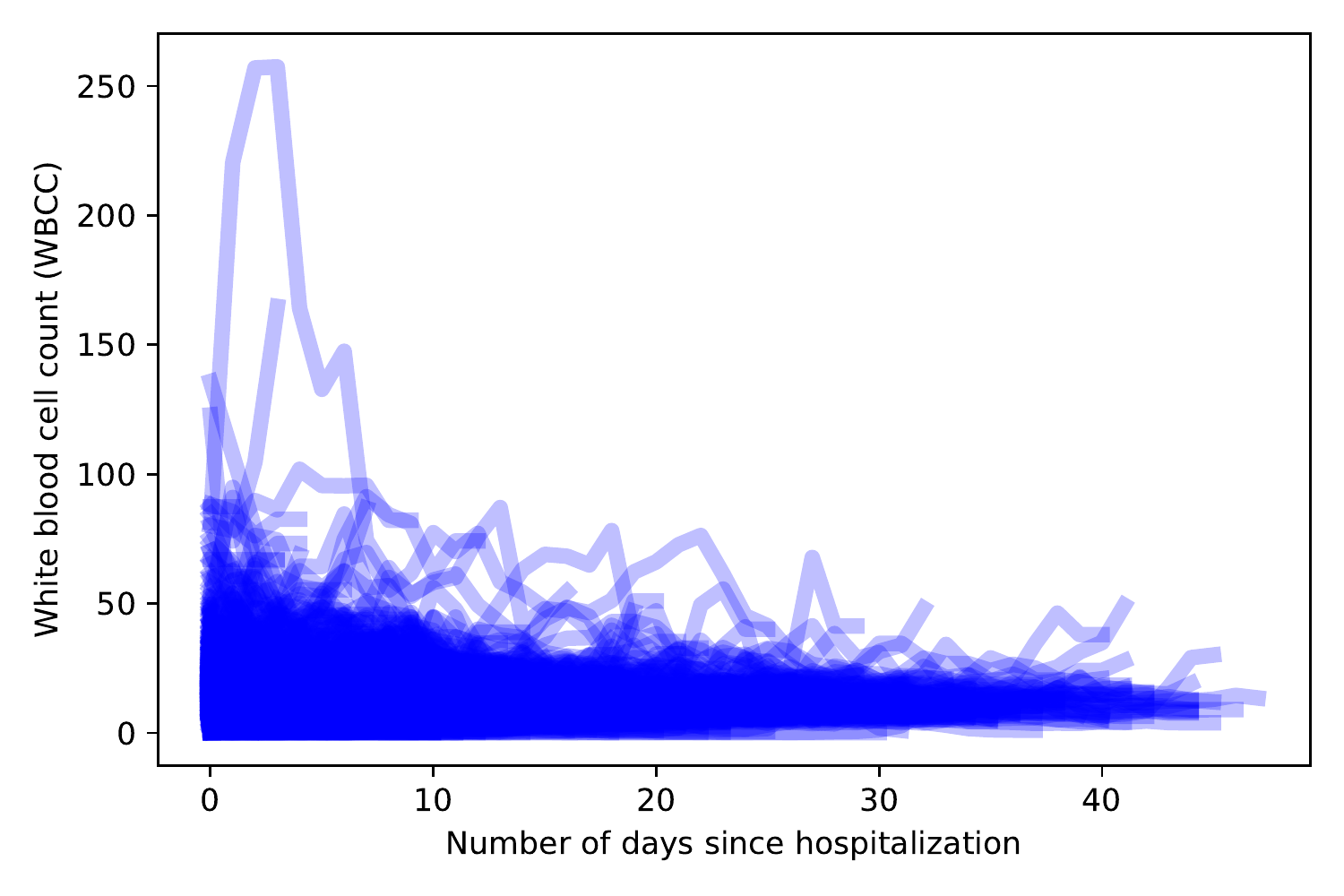}
  %\captionsetup{labelformat=empty, textformat=empty, labelsep=colon}
  \caption{\footnotesize Distribution of white blood cell counts over time.} 
  \vspace{-3mm} 
  \label{FigD2}
\end{figure}

\begin{table}[h]
\centering
{\small
\begin{tabular}{@{}ll@{}}
\toprule
\multicolumn{2}{c}{\textbf{Variable Names}} \\ 
 &   \\ [-2ex]\midrule
Age & Weight \\ [0.25ex]
Temperature & Heart rate \\ [0.25ex]
Systolic blood pressure & Diastolic blood pressure \\ [0.25ex] 
Mean blood pressure & SpO2 \\ [0.25ex]
FiO2 & Respiratory rate \\ [0.25ex] 
Glucose & Bicarbonate (high) \\ [0.25ex] 
Bicarbonate (low) & Creatinine (high) \\ [0.25ex] 
Creatinine (low) & Hematocrit (high)  \\ [0.25ex]
Hematocrit (low) & Hemoglobin (high)  \\ [0.25ex] 
Hemoglobin (low) & Platelet (high)  \\ [0.25ex] 
Platelet (low) & Potassium (high) \\ [0.25ex]
Potassium (low) & Billirubun (high) \\ [0.25ex] 
Billirubun (low) & Weight blood cell (high)\\ [0.25ex] 
Weight blood cell (low) & Antibiotics \\ [0.25ex] 
Norepinephrine & Mechanical ventilator \\ \bottomrule
\end{tabular}
%\captionsetup{labelformat=empty}
\caption{\footnotesize Vital signs and lab tests in MIMIC-III.}}
\label{TableD1}
\end{table}
The follow-up times were varying from one patient to the other, ranging from 4 days to 10 days. This results in a learning problem with variable-length sequences. Clinical care teams decide {\it when} to administer an antibiotic for each patient on a daily basis --- the frequency of antibiotic treatments over time for the entire population is provided in Figure \ref{FigD1}. The patient outcomes under consideration are the white blood
cell count (WBCC) --- a high white blood cell count is associated with severe illness and poor outcome for ICU patients \cite{shuman2012hematologic}. The WBCC averaged for all patients over time are depicted in Figure \ref{FigD2}.

{\bf Clinical setup.} The WBCC levels are medically categories into four levels as follows \cite{waheed2003white}:
\begin{itemize}
\item Leucopenic ($< 4 \times 10^{9}/L$).
\item Normal ($4-10 \times 10^{9}/L$).
\item Leucemoid ($10-25 \times 10^{9}/L$).
\item Exaggerated Leucemoid ($> 25 \times 10^{9}/L$).
\end{itemize}

Antibiotic administration in the ICU aims at keeping the WBCC within the normal range. However, the extent by which an antibiotic affects a patient's WBCC --- if the WBCC reduces significantly and enters the Leucopenic level, then the patient might encounter further health risks. Thus, we need to predict a patient's outcome with confidence intervals; if those intervals overlaps with the Leucopenic and Leucemoid regions, then we should abstain from administering an antibiotic.

We link this setup with our formulation as follows. Each patient's feature sequence $\boldsymbol{x}$ is a collection of 30 clinical variables measured repeatedly over $T$ time steps ($T \leq 10$). We selected the most relevant 5 among these features, and trained an RNN model to predict the next-day WBCC for every patient, and use our procedure to construct confidence intervals on these predictions. 

{\bf Baselines.} We conduct our main experiment with the simple RNN architecture, in addition to LSTMs and GRUs. Further architectures are also explored later in this~Section.~We~compared our method with 5 baselines for predictive uncertainty estimation in RNNs that cover the different modeling categories presented in Section \ref{Sec1}. The modeling approaches under consideration and their respective baselines are:

{\it (a) Bayesian RNNs.} We implemented the Monte Carlo dropout-based Bayesian RNNs (DP-RNNs) proposed in \cite{gal2016theoretically} using the approximate variational inference scheme described in Sections 3 and 4 in \cite{gal2016theoretically}. Dropout is applied to all RNN layers during training and then Monte Carlo dropout is applied for each new test sequence to collect samples of the model predictions. We compute the confidence intervals by estimating the posterior variance of a Gaussian distribution through the Monte Carlo samples and then constructing confidence intervals through the tails of the Gaussian.     

{\it (b) Quantile RNNs.} We implemented an quantile~RNN forecaster similar to the one proposed in \cite{wen2017multi} using the quantile loss: 
\begin{align}
\ell(y, f) = q\,(y-f^{(q)})_{+} + (1-q)\,(f^{(q)}-y)_{+}, \nonumber
\end{align}
where $(.)_+ = \max(0, .)$ and $y^{(q)}$ is the $q$-th quantile of the prediction target. The parameter $q$ is selected so that the learned prediction intervals achieve a coverage probability of $(1-\alpha)$. In order to construct confidence intervals with $\alpha$-level coverage, we train the RNN decoder to predict $f^{(1-\alpha/2)}$ and $f^{(1-\alpha/2)}$ in order to recover the lower and upper limits of the confidence intervals, respectively.

{\it Hyperparameter Tuning and Uncertainty Calibration}

To ensure a fair comparison, we fix the hyperparameters of the underlying RNNs in all baselines to the following: 

\begin{itemize}
\item Number of layers: 1
\item Number of hidden units: 20
\item Learning rate: 0.01
\item Batch size: 150
\item Number of epochs: 10
\item Number of steps: 1000
\end{itemize}

The hyperparameters above where tuned to optimize the RMSE performance of the vanilla RNN model. There was no improvement in the out-of-sample RMSE or discriminative accuracy of all baselines by tweaking these hyperparameters. For the DP-RNN model, we tuned the dropout rate to optimize its RMSE given the values above for the other hyperparameters, the optimal dropout value was 0.45.  

The $(1-\alpha)$ confidence intervals for all baselines were computed as follows:
\begin{itemize}
\item {\bf MQ-RNN:} The confidence intervals are obtained straightforwardly as the interval bounded by the model's quantile outputs $[y^{(\alpha)}, y^{(1-\alpha)}]$. Note that the MQ-RNN is explicitly trained to predict these values.
\item {\bf DP-RNN:} Because DP-based inference approximates a (deep) Gaussian process prosterior, we construct the $(1-\alpha)$ confidence intervals through the tails of the Gaussian, i.e., the confidence interval is given by $[y^{(\alpha)}, y^{(1-\alpha)}]$, where the limits of the interval are constructed as $y^{(q)} = z_c(q) \cdot \sqrt{\mathbb{V}_n[\{y_i\}_i]}$, where $z_c(q)$ is the critical value of the Gaussian at the $q$-th quantile, and $\mathbb{V}_n[\{y_i\}_i]$ is the empirical variance of the Monte Carlo samples of the model. 
\end{itemize}

\end{document}